\documentclass{article}
\newif\ifarxiv
\arxivtrue 
\usepackage{iclr2027_conference,times}

\usepackage{amsmath,amsfonts,bm}

\def\eqref#1{equation~\ref{#1}}

\def\1{\bm{1}}

\DeclareMathAlphabet{\mathsfit}{\encodingdefault}{\sfdefault}{m}{sl}
\SetMathAlphabet{\mathsfit}{bold}{\encodingdefault}{\sfdefault}{bx}{n}

\usepackage{hyperref}
\usepackage{url}

\title{\egohead\hspace{0.2em}LEGO: A Lifting-Free Approach for\texorpdfstring{\\}{ }Exocentric-to-Egocentric Video Generation}

\usepackage{xspace}
\usepackage{fontawesome5}
\usepackage{graphicx}
\usepackage{xcolor}
\usepackage{booktabs}
\usepackage{multirow}
\usepackage{array}
\usepackage{wrapfig}
\usepackage{comment}
\usepackage{tikz}
\usepackage{bbm}

\definecolor{headyellow}{HTML}{F2CD37}
\newcommand{\egohead}{\texorpdfstring{%
  \tikz[baseline=-0.27em, x=1em, y=1em, line width=0.045em, line cap=round]{%
    \filldraw[fill=headyellow, draw=black] (-0.17,0.40) rectangle (0.17,0.56);
    \filldraw[fill=headyellow, draw=black, rounded corners=0.16em] (-0.40,-0.40) rectangle (0.40,0.40);
    \fill[black] (-0.15,0.10) circle[radius=0.055];
    \fill[black] (0.15,0.10) circle[radius=0.055];
    \draw[black, line width=0.05em] (-0.22,-0.10) arc[start angle=215, end angle=325, radius=0.27];
  }}{}}
\newcommand{\method}{\texttt{LEGO}\xspace}

\newcommand{\quoted}[1]{\textcolor{black!55}{#1}} 

\definecolor{darker}{rgb}{0.15,0.45,0.4}
\hypersetup{colorlinks=true, urlcolor=darker, citecolor=darker, linkcolor=darker}
\author{Suhwan Cho\textsuperscript{*}\quad Yonwoo Choi\textsuperscript{*}\quad Soongjin Kim\textsuperscript{*} \quad Jicheol Park \quad Taegyu Lim\vspace{0.2cm}\\
GenGenAI\vspace{0.2cm}\\
\makebox[1.15em][c]{\raisebox{-0.15em}{\faGithub}}\ \textbf{\method}:\ {\small\url{https://github.com/suhwan-cho/lego}}\\[0.1em]
\makebox[1.15em][c]{\raisebox{-0.2em}{\includegraphics[height=1.05em]{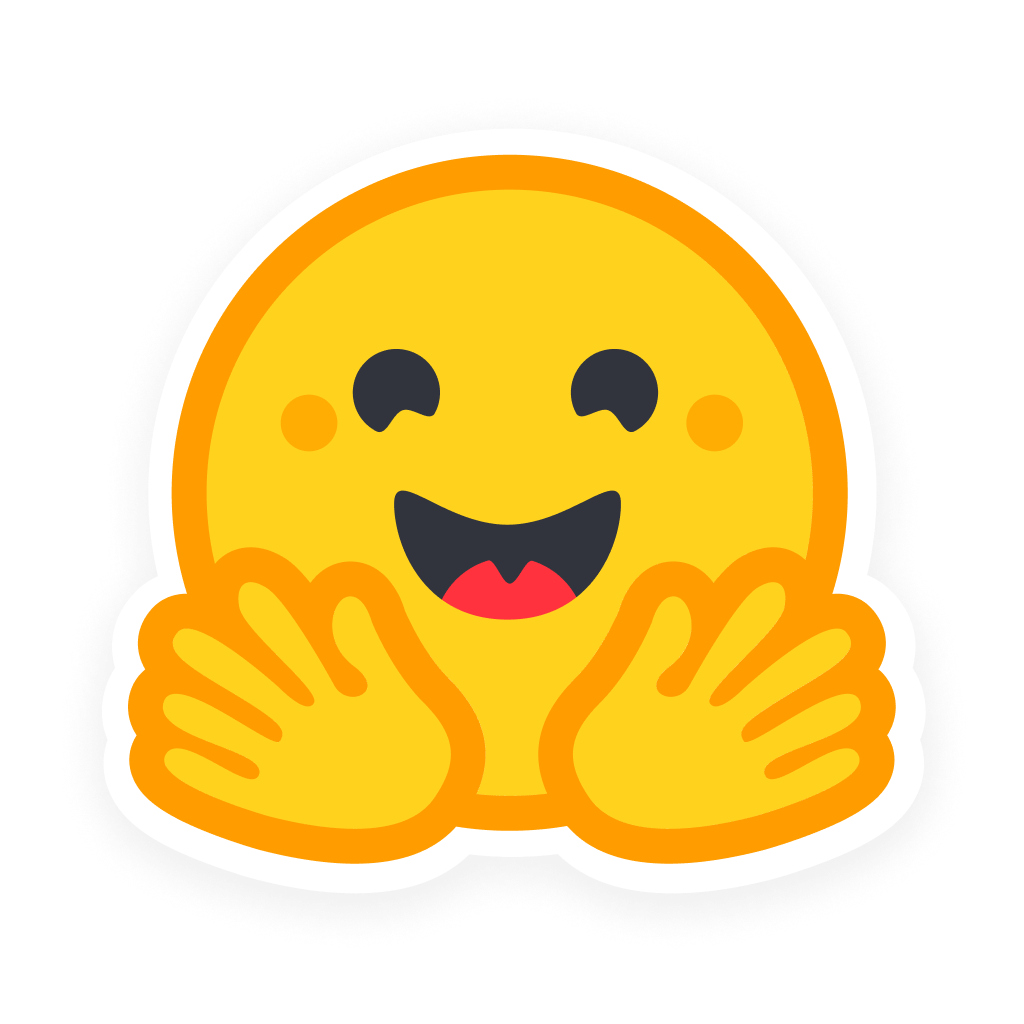}}}\ \textbf{\method}:\ {\small\url{https://huggingface.co/suhwan-cho/lego}}%
}

\iclrfinalcopy 
\ifarxiv\iclrfinalcopy\fi

\begin{document}
\maketitle
\ifarxiv\lhead{}\renewcommand{\headrulewidth}{0pt}\fi
\ificlrfinal
\begingroup
\renewcommand\thefootnote{}%
\footnotetext{\textsuperscript{*}Denotes equal contribution.}%
\addtocounter{footnote}{-1}%
\endgroup
\fi

\begin{figure}[!ht]
\vspace{-10pt} 
\centering
\includegraphics[width=\linewidth]{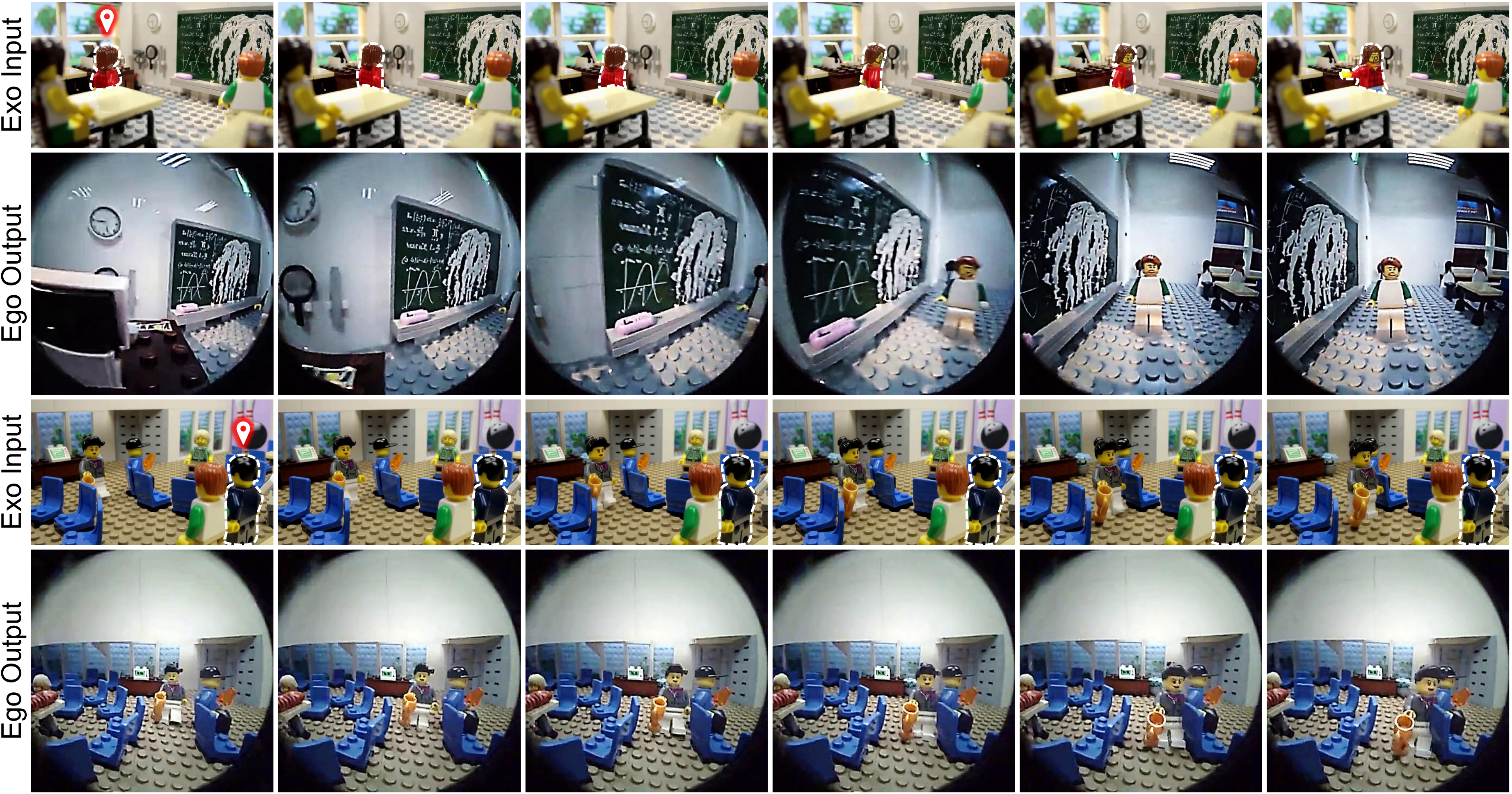}
\vspace{-20pt}
\caption{Egocentric videos generated by \method from real-world exocentric footage, viewed from the perspective of the LEGO minifigure marked by the dotted outline.}
\vspace{15pt}
\label{fig:teaser}
\end{figure}

\begin{abstract}
Generating an egocentric video from a single exocentric recording is a challenging case of novel view synthesis, as the two cameras share little overlap and much of the target view is unobserved. Current state-of-the-art methods reconstruct the scene explicitly by estimating depth, lifting the video into a point cloud, and re-rendering it from the egocentric camera to condition a video diffusion model. This deterministic mapping assigns each pixel to a single reprojected location, which preserves texture but translates depth errors into misplaced content. We ask what a video diffusion model should receive as its condition and propose a lifting-free answer: a learned view synthesizer, an LVSM-style transformer fine-tuned to render the egocentric view directly without depth, point clouds, or reprojection, resolving cross-view correspondence internally. In contrast, its probabilistic mapping averages each region over candidate source locations according to a learned correspondence distribution, preserving structure while fine texture is averaged away. We argue that this trade-off suits a diffusion generator, whose denoising training excels at restoring detail, so an effective condition should prioritize structural alignment over sharpness. This distribution's concentration also yields a per-region confidence, used both to mask low-confidence regions and to guide the generator toward high-confidence areas during early layout-forming denoising steps. Our approach consistently outperforms the state-of-the-art explicit pipeline and generalizes to other datasets without retraining. The synthesizer thus supplies view structure, and the diffusion model its detail.
\end{abstract}

\section{Introduction}
We study exocentric-to-egocentric video generation: given a single third-person recording of a scene and a target egocentric camera trajectory, the goal is to synthesize the video a camera following this trajectory would capture. The task matters wherever first-person data is scarce and third-person footage abundant \citep{grauman2024egoexo4d}, and it is demanding: the two views share little content, so much of the target must be synthesized rather than copied. Existing methods condition implicitly on exocentric features \citep{luo2024put,cheng2024fourdiff}; assume privileged inputs such as surrounding cameras \citep{liu2024exocentric}, rich multimodal observations \citep{park2025egoworld}, or a ground-truth egocentric first frame \citep{xu2025egoexogen}; or reconstruct the scene explicitly. The current state of the art \citep{kang2026egox} takes the explicit route, conditioning a video diffusion model \citep{wan2025wan} on a point-cloud render obtained by lifting the exocentric video with monocular depth and replaying it from the egocentric trajectory. This route inherits the errors of the estimated depth: depth errors misplace the content of the point-cloud render, sharply and without any signal of where it is wrong.

We present \method (\textbf{L}ifting-free exocentric-to-\textbf{EGO}centric video generation; Fig.~\ref{fig:method}), a lifting-free alternative that replaces the reconstruction pipeline with a learned view synthesizer that renders the egocentric view directly. An LVSM-style transformer \citep{jin2025lvsm} is fine-tuned to map the exocentric frame and the target camera rays to the egocentric image, so cross-view correspondence is learned from the rendering objective alone. The frozen synthesizer's render replaces the point-cloud render in the generator's conditioning input; no depth estimation, point cloud, or reprojection is involved.

The design follows from the paper's central question: what should a video diffusion model be given as its condition? The two renders fail in opposite ways. The point-cloud render is deterministic: each pixel is assigned a single reprojected location, so its errors are misplaced content. The synthesizer's render is probabilistic: each region averages over candidate source locations according to a learned correspondence distribution, so its errors are blur. These failures are not equally costly: blur is the corruption that denoising training teaches the generator to undo, and diffusion models routinely recover sharp detail from blurred inputs \citep{saharia2023sr3,ho2022cascaded,meng2022sdedit}, whereas rough or misplaced conditions have needed dedicated mechanisms to relax them \citep{liu2024smartcontrol,seo2024genwarp}. We therefore argue that a condition should be judged by alignment first and sharpness second, since the generator supplies the second far more readily than the first: the condition should say where things are, and the generator will decide how they look. The correspondence distribution supplies a further signal for this: its concentration at each region is a confidence, and we mask from the conditioning those regions where it is low.
 
Alignment first, detail second also describes the sampling trajectory itself: layout is decided in the early, high-noise steps and detail emerges later, so structural guidance is most useful early. We therefore reuse the same confidence at inference, guiding the early denoising steps toward the render where it is high and leaving the rest unconstrained. This requires no additional training; we call it Adaptive Rendering Consistency (ARC, Section~\ref{sec:arc}).

On Ego-Exo4D, the synthesizer's render outperforms the point-cloud render as a condition, ARC at inference alone outperforms the depth-derived attention bias with which prior work trains the generator, and the full system outperforms the state-of-the-art explicit pipeline on seen and unseen scenes; the same weights also generalize to EgoHumans and Nymeria without retraining.

Our contributions are: \textbf{(1)} \method, a lifting-free exocentric-to-egocentric video generation framework that conditions a video diffusion model on the confidence-gated render of a learned view synthesizer instead of a point-cloud reconstruction; \textbf{(2)} evidence that a structurally aligned but soft condition serves the generator better than a sharp but misplaced one; \textbf{(3)} ARC, training-free guidance that reuses the synthesizer's confidence to steer the early denoising steps toward its render; and \textbf{(4)} state-of-the-art results on Ego-Exo4D and generalization to EgoHumans and Nymeria without retraining.

\section{Related Work}

\vspace{-1mm}
\paragraph{Exocentric-to-egocentric video generation.} Early systems relied on privileged inputs: surrounding cameras \citep{liu2024exocentric}, rich multimodal exocentric observations \citep{park2025egoworld}, or the ground-truth egocentric first frame \citep{xu2025egoexogen}. Others condition implicitly on exocentric features \citep{luo2024put,cheng2024fourdiff}. The state of the art, EgoX \citep{kang2026egox}, takes a single exocentric video and reconstructs the scene explicitly: depth from a video pose engine \citep{huang2025vipe} lifts the frames into a point cloud, which is re-rendered along the egocentric trajectory to condition a video diffusion model \citep{wan2025wan}, and the generator is trained with a depth-derived attention bias. Concurrent and adjacent work extends this explicit line \citep{gu2026e3c,song2026worldwander,mahdi2026synchrony}, reverses the direction or transfers exocentric knowledge \citep{luo2024intention,xiu2026egotwin,zhang2026exo2ego}, or pairs the two views for recognition \citep{sigurdsson2018actor,li2021egoexo}. \method keeps the single-video setting and diffusion backbone of EgoX, but replaces the reconstruction with a learned view synthesizer and the training-time attention bias with training-free guidance.

\vspace{-1mm}
\paragraph{Camera-controlled video generation.} Camera poses can steer video diffusion directly, through extrinsics or Pl\"ucker embeddings injected into the backbone \citep{he2025cameractrl,wang2024motionctrl,bahmani2025ac3d} or by re-rendering a video along a new trajectory \citep{bai2025recammaster}. For larger viewpoint changes, systems lift the video into explicit 3D and condition generation on a render of the target trajectory \citep{yu2024viewcrafter,yu2025trajectorycrafter,ren2025gen3c,huang2025voyager}. EgoX follows this recipe in a setting where the view change is extreme and estimated depth is least reliable; \method takes the same pose inputs but is lifting-free.

\vspace{-1mm}
\paragraph{Learned view synthesis and emergent correspondence.} Novel view synthesis has moved from per-scene optimization to feed-forward models that predict radiance fields, 3D Gaussians, or triplanes in a single pass \citep{yu2021pixelnerf,liu2023zero,hong2024lrm,charatan2024pixelsplat,chen2024mvsplat}; LVSM \citep{jin2025lvsm} maps posed input tokens to target-ray tokens with a plain transformer. Rendering supervision is known to teach geometry: photometric view synthesis losses made monocular depth learnable without ground truth \citep{garg2016unsupervised,zhou2017unsupervised}. Self-supervised and diffusion features likewise acquire correspondence they were never supervised for \citep{caron2021emerging,amir2021deep,tang2023emergent,zhang2023tale,luo2023diffusion}. We rely on the same effect in a synthesizer fine-tuned for the exocentric-to-egocentric view change: its render serves as the generator's condition, and the concentration of its correspondence distribution as a per-region confidence (Section~\ref{sec:cond}).

\vspace{-1mm}
\paragraph{What a diffusion model needs from its condition.} Conditional diffusion models recover fine detail from coarse inputs: super-resolution and cascaded generation upsample low-resolution images, and blurring the condition during training helps rather than hurts \citep{saharia2023sr3,ho2022cascaded}; SDEdit synthesizes realistic images from stroke layouts \citep{meng2022sdedit}; GeNVS produces sharp novel views from a volume-rendered feature image \citep{chan2023genvs}; and blur can serve as the forward corruption a diffusion model learns to invert \citep{rissanen2023heat,hoogeboom2023blurring}, with coarse structure decided at high noise and detail at low noise \citep{choi2022p2,rissanen2023heat}. Conversely, precise conditions are neither required nor always safe: layout-level depth suffices for depth-conditioned control \citep{bhat2024loosecontrol}, control must be relaxed where a rough condition is wrong \citep{liu2024smartcontrol}, and generation from a depth-based reprojection must learn where to trust it \citep{seo2024genwarp}. We read these results as favoring a condition that is aligned before it is sharp.

\vspace{-1mm}
\paragraph{Training-free guidance in diffusion models.} Pretrained diffusion models can be steered at sampling time: ILVR \citep{choi2021ilvr}, RePaint \citep{lugmayr2022repaint}, and Blended Diffusion \citep{avrahami2022blended} blend a reference into the intermediate sample; DDRM \citep{kawar2022ddrm} and DDNM \citep{wang2023ddnm} project each step onto measurement-consistent solutions; DPS \citep{chung2023dps}, MCG \citep{chung2022mcg}, and FreeDoM \citep{yu2023freedom} apply guidance gradients. All assume the reliable content and its mask are given; ARC obtains both from the synthesizer and acts only in the early steps.

\section{Method}

\begin{figure}[t]
\centering
\includegraphics[width=\linewidth]{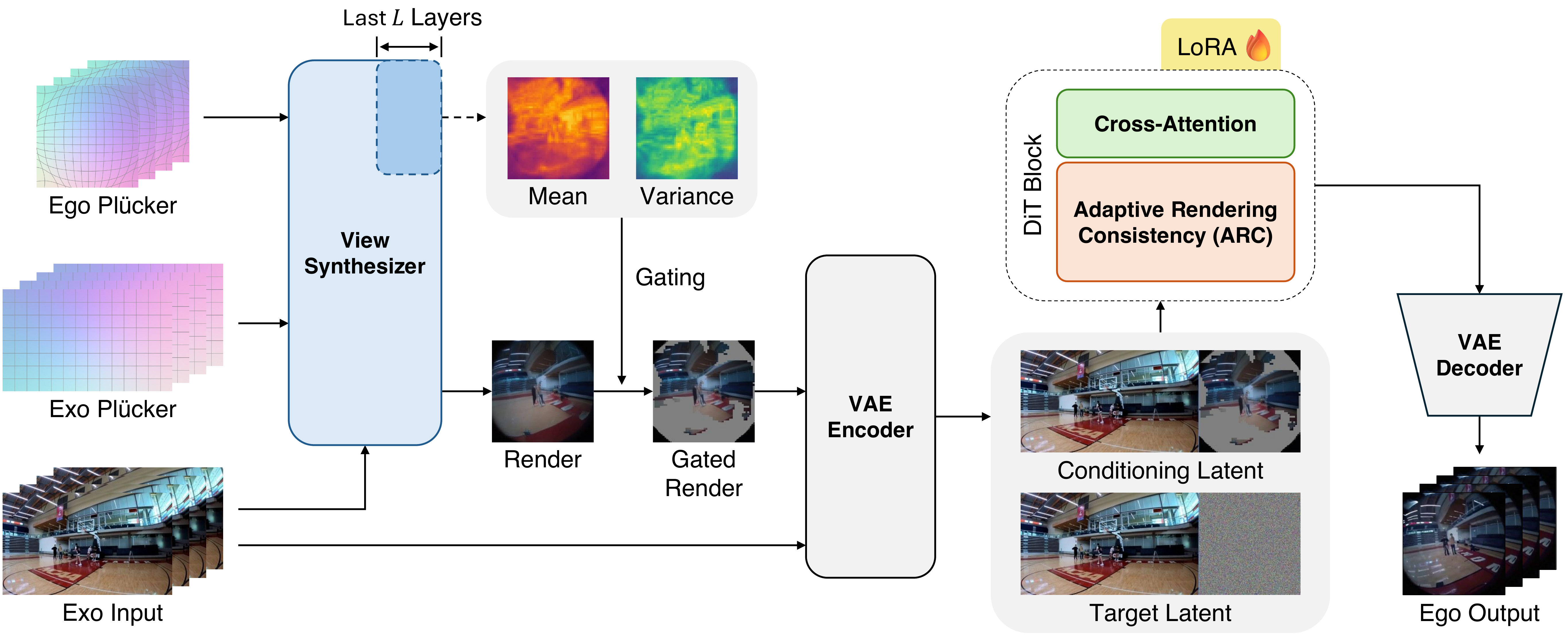}
\caption{Overview of \method. The frozen synthesizer $F$ renders the egocentric view once per latent frame and exposes a confidence map. The render, gated to gray where confidence is low, fills the egocentric half of the conditioning latent; the confidence also drives ARC during the early denoising steps. The generator is unchanged.}
\label{fig:method}
\end{figure}

We follow the input setting of EgoX~\citep{kang2026egox}: a single exocentric video with its camera calibration, and a target egocentric camera trajectory along which the video is to be generated. All experiments use the same image-to-video diffusion backbone and canvas formulation as EgoX, in which the exocentric and egocentric views are concatenated along the width and the egocentric half of the conditioning latent carries the condition for the target view. Prior work fills this conditioning input with a point-cloud render and, in addition, trains the generator with a depth-derived bias on the cross-view attention. \method fills it with the render of a frozen view synthesizer (Section~\ref{sec:prior}), gates the render by the confidence of the synthesizer's correspondence distribution (Section~\ref{sec:cond}), and uses the same confidence to guide the early denoising steps toward the render (Section~\ref{sec:arc}). The pipeline is lifting-free: no depth is estimated and no point cloud is built at any stage. We do not modify the generator. We change the condition it receives and how it is sampled (Fig.~\ref{fig:method}).

\subsection{Learned View Synthesizer}\label{sec:prior}

We train a view synthesizer $F$, an LVSM-style transformer \citep{jin2025lvsm}, to render the egocentric view directly. Geometry enters $F$ only through per-pixel ray maps: a calibrated camera assigns each pixel $\mathbf{u}$ the Pl\"ucker embedding $\mathbf{r}(\mathbf{u})=(\mathbf{d}(\mathbf{u}),\,\mathbf{o}\times\mathbf{d}(\mathbf{u}))\in\mathbb{R}^6$ of its viewing ray, with unit direction $\mathbf{d}$ and camera center $\mathbf{o}$. All poses are expressed in the exocentric camera frame, so the moment $\mathbf{o}\times\mathbf{d}(\mathbf{u})$ of each egocentric ray carries the metric baseline between the two cameras. For frame $t$, $F$ takes the exocentric frame and the two ray maps and predicts the egocentric frame,
\begin{equation}\label{eq:render}
\hat{I}_t^{\mathrm{ego}}
=
F\!\left(I_t^{\mathrm{exo}},\ \mathbf{r}^{\mathrm{exo}},\ \mathbf{r}_t^{\mathrm{ego}}\right),
\end{equation}
and is trained with an $\ell_2$ and a perceptual loss against the ground-truth egocentric frame. The target frame is the only supervision; no depth or correspondence labels are used.

\vspace{-1mm}
\paragraph{Correspondence distribution.} Inside $F$, each egocentric query $\mathbf{u}$ attends over exocentric locations $\mathbf{v}$ with normalized weights $w(\mathbf{u}\!\to\!\mathbf{v})$. Averaging these weights over heads and over the last $L$ layers gives, for every target location, a distribution over source locations,
\begin{equation}\label{eq:corr}
p_{\mathbf{u}}(\mathbf{v})
=
\operatorname*{mean}_{\text{heads,\ last }L\text{ layers}}
w(\mathbf{u}\!\to\!\mathbf{v}),
\end{equation}
which we call the correspondence distribution. At the image level, it makes the difference between the two pipelines explicit:
\begin{equation}\label{eq:transport}
\underbrace{
\hat{I}^{\mathrm{ego}}(\mathbf{u})
\approx
\sum_{\mathbf{v}}
p_{\mathbf{u}}(\mathbf{v})\,
I^{\mathrm{exo}}(\mathbf{v})
}_{\text{learned (probabilistic)}}
\qquad\qquad
\underbrace{
\hat{I}^{\mathrm{ego}}_{\pi}(\mathbf{u})
=
I^{\mathrm{exo}}\!\left(\pi_d(\mathbf{u})\right)
}_{\text{explicit (deterministic)}},
\end{equation}
where $\pi_d(\mathbf{u})$ denotes the exocentric source location corresponding to the ego-view pixel $\mathbf{u}$, as determined by geometric reprojection using the estimated depth $d$. The approximation on the left reflects that $F$ mixes features across layers rather than averaging pixels directly.

\begin{figure}[t]
\centering
\includegraphics[width=\linewidth]{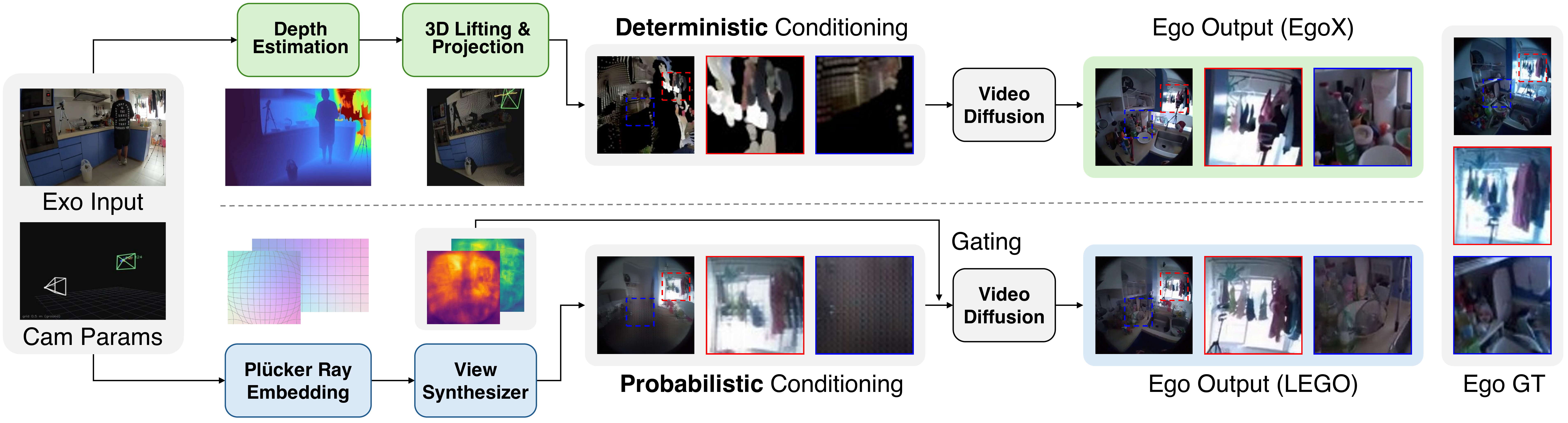}
\caption{Condition generation. EgoX estimates depth, lifts the exocentric video into a point cloud, and re-renders it along the egocentric trajectory; \method renders the egocentric view directly with a frozen synthesizer.}
\label{fig:cond}
\end{figure}

\vspace{-1mm}
\paragraph{Two kinds of error.} The two mappings fail differently. The explicit mapping assigns each target pixel a single source pixel, which keeps texture sharp but translates a depth error directly into misplaced content; in statistical terms, its error is a bias. The learned mapping averages over the candidate sources in $p_{\mathbf{u}}$, which blurs texture wherever the distribution is spread out but keeps content close to its correct place; its error is a variance. Denoising training makes the generator well suited to restoring detail \citep{saharia2023sr3,ho2022cascaded,chan2023genvs}, so blur is the failure it is best equipped to correct, whereas conditions that are sharp but misplaced have required dedicated mechanisms to relax or override them \citep{liu2024smartcontrol,seo2024genwarp}. This is why we condition on $F$'s render rather than on the point-cloud render; Section~\ref{sec:exp} tests the choice, and Fig.~\ref{fig:cond} contrasts the two pipelines.

\vspace{-1mm}
\paragraph{Adapting to the egocentric camera.}\label{sec:method-camera} The published LVSM checkpoint, which we call the untuned synthesizer, is trained to interpolate between nearby pinhole views, whereas our target is an extrapolation to a fisheye headset camera that shares little overlap with the source. Because every camera-specific quantity enters $F$ through the ray map, no architectural change is needed: we invert the calibrated fisheye projection per pixel to obtain the target rays, and fine-tune all of its weights on exocentric-egocentric pairs, supervised by the raw fisheye recordings. Supervising on real recordings also teaches $F$ where the lens does not see, which the ray map alone cannot express (Appendix~\ref{sec:spill}). The fine-tuning is cheap, 10{,}000 steps or about 3.5 hours on four NVIDIA  H200 GPUs, and $F$ is frozen afterwards.

\subsection{Confidence-Gated Conditioning}\label{sec:cond}

The render of $F$ replaces the point-cloud render in the egocentric half of the conditioning latent. The two differ in one further respect: the point-cloud render leaves a hole wherever the geometry offers no support, whereas the probabilistic mapping fills every location, including those with no exocentric evidence, and so cannot abstain on its own. The concentration of $p_{\mathbf{u}}$ supplies the missing signal. We define the confidence of a target location as the probability mass of its $k$ most likely sources, and set locations below a threshold to neutral gray:
\begin{equation}\label{eq:conf}
c(\mathbf{u})
=
\sum_{\mathbf{v}\in \operatorname{TopK}_k(p_{\mathbf{u}})}
p_{\mathbf{u}}(\mathbf{v}),
\qquad
C(\mathbf{u})
=
\begin{cases}
\hat{I}^{\mathrm{ego}}(\mathbf{u}),
& c(\mathbf{u}) \ge \tau,\\
\gamma,
& \text{otherwise},
\end{cases}
\end{equation}
where $c(\mathbf{u})\in[0,1]$, $\tau$ is the gating threshold, and $\gamma$ is a neutral gray that the VAE maps near zero. The gated render $C$ is encoded into the generator's latent space and used as the conditioning input. Since $F$ is frozen, its renders and confidence maps are precomputed once per clip, and the generator is otherwise unchanged, so the comparison in Section~\ref{sec:exp} isolates the effect of the conditioning input alone.

\subsection{Adaptive Rendering Consistency}\label{sec:arc}

The gated render enters the generator as its conditioning input. Prior work additionally steers the generator with a depth-derived bias on cross-view attention, applied throughout training. We instead act on the denoising trajectory at inference time: during the early steps, in which the layout of the sample is determined \citep{choi2022p2,rissanen2023heat}, the generator's current estimate of the clean video is moved toward the render in proportion to the confidence, and the remaining steps are left unconstrained. We call this \emph{Adaptive Rendering Consistency} (ARC).

The generator is trained with flow matching: a noisy latent
$z_\sigma=(1-\sigma)\,z_0+\sigma\,\epsilon$ is formed at noise level
$\sigma\in[0,1]$, the transformer predicts the velocity $v\approx\epsilon-z_0$,
and sampling follows a decreasing schedule $\sigma_0=1>\sigma_1>\dots>\sigma_n=0$.
At step $i$ the solver moves from $z_i$ to $z_{i+1}$, and the same velocity
implies an estimate of the clean video at the current step:
\begin{equation}\label{eq:flow}
z_{i+1} = z_i+(\sigma_{i+1}-\sigma_i)\,v,
\qquad
\hat{z}_0 = z_i-\sigma_i\,v .
\end{equation}
Since $\sigma_{i+1}<\sigma_i$, each step moves the sample a fixed fraction of the
way toward $\hat{z}_0$, so the clean estimate sets the direction of every step.
ARC acts there: during the first $hn$ steps it replaces the egocentric part of
the clean estimate by a confidence-weighted blend with the render, and passes the
corresponding velocity to the solver:
\begin{equation}\label{eq:arc}
\hat{z}_0'^{\,\mathrm{ego}} = \bigl(1-\mathbf{c}^2\bigr)\odot\hat{z}_0^{\mathrm{ego}}+\mathbf{c}^2\odot E,
\qquad
v'^{\,\mathrm{ego}} = \frac{z_i^{\mathrm{ego}}-\hat{z}_0'^{\,\mathrm{ego}}}{\sigma_i}
= v^{\mathrm{ego}}-\frac{\mathbf{c}^2}{\sigma_i}\odot\bigl(E-\hat{z}_0^{\mathrm{ego}}\bigr),
\end{equation}
where $E$ is the ungated render encoded into latent space and $\mathbf{c}$ is the confidence map of Eq.~\ref{eq:conf}, resampled to the latent grid. The second equality follows from $\hat{z}_0=z_i-\sigma_i v$. The denoising step is then taken exactly as in Eq.~\ref{eq:flow} with $v'$ in place of $v$,
\begin{equation}\label{eq:step}
z_{i+1}^{\mathrm{ego}} = z_i^{\mathrm{ego}}+(\sigma_{i+1}-\sigma_i)\,v'^{\,\mathrm{ego}},
\end{equation}
so the sample is drawn toward the blended estimate instead of toward $\hat{z}_0$ alone: strongly toward the render where confidence is high, and almost unchanged where it is low. The correction is deterministic and adds no noise, and the solver itself is untouched; in practice we use a second-order multistep solver \citep{zhao2023unipc}, which consumes $v'$ through the same clean estimate. After the first $hn$ steps the trajectory is unconstrained, so the render shapes the layout in the early, high-noise steps while the remaining steps add detail that the render does not contain.

We weight by $\mathbf{c}^2$ rather than $\mathbf{c}$: the squared weight lowers the correction in regions of intermediate confidence, where the render is often blurred, while leaving high-confidence regions near full strength. The ungated render is used because $\mathbf{c}^2$ already suppresses low-confidence regions continuously. The noise level $\sigma_i$ in Eq.~\ref{eq:arc} is lower-bounded by a small constant for numerical stability.

\begin{table}[t]
\caption{Comparison on Ego-Exo4D. Unmarked rows are quoted from \citet{kang2026egox}; $^\dagger$released EgoX checkpoint rerun under our scoring stack. TF, MS, and DD are VBench temporal flickering, motion smoothness, and dynamic degree. Best in bold.}
\label{tab:main}
\vspace{1mm}
\centering
\footnotesize
\setlength{\tabcolsep}{0pt}
\begin{tabular}{p{0.03\linewidth} p{0.20\linewidth} *{8}{>{\centering\arraybackslash}p{0.095\linewidth}}}
\toprule
& & \multicolumn{4}{c}{Image Criteria} & \multicolumn{4}{c}{Video Criteria} \\
\cmidrule(lr){3-6} \cmidrule(lr){7-10}
& Method & PSNR$\uparrow$ & SSIM$\uparrow$ & LPIPS$\downarrow$ & CLIP-I$\uparrow$ & FVD$\downarrow$ & TF$\uparrow$ & MS$\uparrow$ & DD$\uparrow$ \\
\midrule
\multirow{7}{*}{\rotatebox[origin=c]{90}{\textit{Seen}}\hspace{4pt}}
& Exo2Ego-V & 14.53 & 0.384 & 0.569 & 0.774 & 622.47 & 0.960 & 0.966 & 0.985 \\
& TrajectoryCrafter & 13.05 & 0.375 & 0.606 & 0.780 & 546.09 & 0.960 & 0.980 & 0.947 \\
& Wan Fun Control & 12.25 & 0.463 & 0.617 & 0.810 & 595.07 & 0.968 & 0.980 & 0.901 \\
& Wan VACE & 12.95 & 0.413 & 0.626 & 0.829 & 508.69 & \textbf{0.989} & \textbf{0.994} & 0.673 \\
& EgoX & 16.05 & 0.556 & 0.498 & 0.896 & 184.47 & 0.977 & 0.990 & 0.974 \\
& EgoX$^\dagger$ & 15.57 & 0.556 & 0.466 & 0.903 & 197.95 & 0.985 & 0.993 & \textbf{0.990} \\
& \method & \textbf{20.28} & \textbf{0.689} & \textbf{0.310} & \textbf{0.925} & \textbf{155.56} & \textbf{0.989} & \textbf{0.994} & 0.987 \\
\midrule
\multirow{7}{*}{\rotatebox[origin=c]{90}{\textit{Unseen}}\hspace{4pt}}
& Exo2Ego-V & 12.70 & 0.439 & 0.597 & 0.679 & 1283.50 & 0.971 & 0.976 & 0.978 \\
& TrajectoryCrafter & 12.24 & 0.297 & 0.619 & 0.778 & 821.71 & 0.966 & 0.984 & 0.944 \\
& Wan Fun Control & 13.59 & 0.439 & 0.604 & 0.799 & 968.78 & 0.971 & 0.985 & 0.944 \\
& Wan VACE & 12.17 & 0.345 & 0.638 & 0.820 & 1045.45 & \textbf{0.995} & \textbf{0.996} & 0.427 \\
& EgoX & 14.38 & 0.457 & 0.552 & 0.877 & 440.64 & 0.981 & 0.992 & 0.989 \\
& EgoX$^\dagger$ & 13.53 & 0.437 & 0.542 & 0.886 & 455.06 & 0.986 & 0.994 & 0.990 \\
& \method & \textbf{15.91} & \textbf{0.505} & \textbf{0.504} & \textbf{0.897} & \textbf{404.77} & 0.990 & 0.995 & \textbf{1.000} \\
\bottomrule
\end{tabular}
\end{table}

\section{Experiments}\label{sec:exp}
\vspace{-1mm}
\paragraph{Experimental setup.} We train and evaluate on Ego-Exo4D \citep{grauman2024egoexo4d} following the protocol of EgoX \citep{kang2026egox}. The benchmark has two test splits: \emph{seen}, whose scenes appear in the training set, and \emph{unseen}, whose scenes do not. PSNR, SSIM, LPIPS, CLIP-I, and FVD are computed on the generated egocentric frames, and the VBench scores for temporal flickering (TF), motion smoothness (MS), and dynamic degree (DD) on the full exocentric--egocentric canvas. We compare with the numbers reported in \citet{kang2026egox} and with the released EgoX checkpoint rerun under our evaluation code, denoted EgoX$^\dagger$.
For a fair comparison, all ablations use the same backbone, LoRA rank, training schedule, random seed, and camera trajectories. They differ only in the conditioning input and in how denoising is guided. Implementation details, the evaluation protocol, and object-level metrics are given in Appendices~\ref{app:impl}, \ref{app:protocol}, and \ref{app:object}, respectively.

\begin{table}[t]
\caption{Ablation under one recipe. \emph{Top}: what fills the conditioning input (exocentric tokens hidden). \emph{Middle}: cumulative additions up to the full conditioning. \emph{Bottom}: how denoising is driven: GGA applied at training and inference, ARC at inference only, and ARC anchored to the point-cloud render with its hole mask as binary confidence. ARC (synthesizer render) is the \method row of Table~\ref{tab:main}. Best in bold.}
\label{tab:factorial}
\vspace{1mm}
\centering
\footnotesize
\setlength{\tabcolsep}{0pt}
\begin{tabular}{p{0.28\linewidth} *{8}{>{\centering\arraybackslash}p{0.09\linewidth}}}
\toprule
& \multicolumn{4}{c}{Seen} & \multicolumn{4}{c}{Unseen} \\
\cmidrule(lr){2-5} \cmidrule(lr){6-9}
Arm & PSNR$\uparrow$ & LPIPS$\downarrow$ & CLIP-I$\uparrow$ & FVD$\downarrow$ & PSNR$\uparrow$ & LPIPS$\downarrow$ & CLIP-I$\uparrow$ & FVD$\downarrow$ \\
\midrule
\multicolumn{9}{@{}l}{\textit{Conditioning input: what fills it}} \\
\quad No condition & 12.30 & 0.591 & 0.862 & 310.00 & 11.38 & 0.640 & 0.838 & 695.44 \\
\quad Point-cloud render & 15.59 & 0.478 & 0.895 & 195.80 & 13.42 & 0.595 & 0.853 & 549.77 \\
\quad Untuned synthesizer render & 15.71 & 0.503 & 0.888 & 212.94 & 13.56 & 0.596 & 0.848 & 560.51 \\
\quad Synthesizer render & 18.24 & 0.364 & 0.904 & 183.40 & 14.35 & 0.567 & 0.860 & 581.90 \\
\midrule
\multicolumn{9}{@{}l}{\textit{Completing the configuration (cumulative)}} \\
\quad\quad $+$ confidence gating & 18.27 & 0.375 & 0.904 & 166.86 & 14.40 & 0.566 & 0.868 & 502.47 \\
\quad\quad $+$ exocentric tokens & 18.91 & 0.340 & 0.921 & 159.04 & 14.98 & 0.515 & \textbf{0.897} & 415.08 \\
\midrule
\multicolumn{9}{@{}l}{\textit{Denoising process: how it is driven}} \\
\quad\quad GGA & 18.62 & 0.356 & 0.915 & 194.98 & 15.20 & 0.525 & 0.896 & 429.42 \\
\quad\quad ARC (point-cloud render) & 13.96 & 0.586 & 0.801 & 735.50 & 13.72 & 0.618 & 0.795 & 812.88 \\
\quad\quad ARC (synthesizer render) & \textbf{20.28} & \textbf{0.310} & \textbf{0.925} & \textbf{155.56} & \textbf{15.91} & \textbf{0.504} & \textbf{0.897} & \textbf{404.77} \\
\bottomrule
\end{tabular}
\end{table}
%

\begin{figure}[t]
\centering
\includegraphics[width=\linewidth]{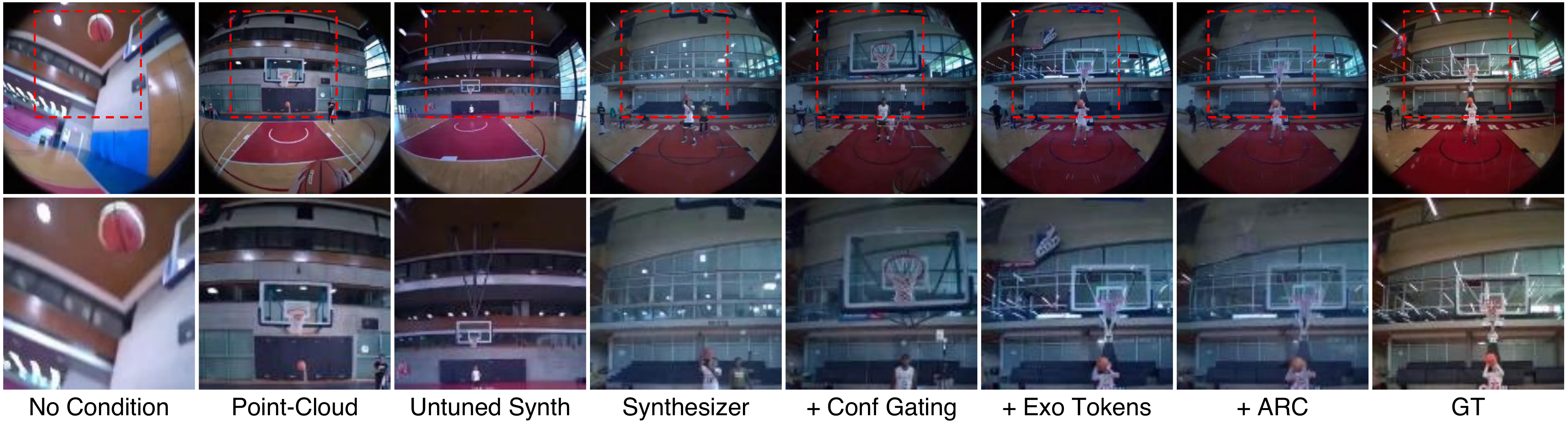}
\vspace{-20pt}
\caption{Qualitative ablation on one unseen clip, following the rows of Table~\ref{tab:factorial}. The second row shows enlarged crops of the corresponding regions in the first row. }
\label{fig:condresults}
\end{figure}

\vspace{-1mm}
\paragraph{Quantitative evaluation.} Table~\ref{tab:main} reports the comparison with prior methods on both splits. \method outperforms EgoX on all image criteria and FVD, with the largest gains on PSNR and LPIPS. The other baselines perform substantially worse than both. Temporal flickering and motion smoothness are near saturation for all methods, and dynamic degree is comparable.

\vspace{-1mm}
\paragraph{Ablation Study.} Table~\ref{tab:factorial} ablates the conditioning input and the denoising guidance under the same training recipe, and Fig.~\ref{fig:condresults} shows the corresponding generations.

\vspace{-1mm}
\emph{Conditioning input.} We first compare what fills the conditioning input, with the exocentric tokens hidden. The point-cloud render improves over an empty condition, and the fine-tuned synthesizer render improves further on both splits. The untuned synthesizer render performs on par with the point-cloud render, which shows that the gain comes from adapting $F$ to the egocentric camera (Section~\ref{sec:method-camera}) rather than from its architecture.

\vspace{-1mm}
\emph{Gating and exocentric tokens.} Confidence gating discards part of the rendered region, yet it lowers FVD on both splits and leaves the other metrics unchanged. Restoring the exocentric tokens brings a further gain on all metrics, as the model can also refer to the source view directly.

\vspace{-1mm}
\emph{Denoising guidance.} On top of the full conditioning, ARC improves all metrics without any retraining and yields the final \method model in Table~\ref{tab:main}. GGA, the depth-guided attention bias of EgoX, applied at training and inference as in EgoX, does not improve over the full conditioning. Applying ARC to the point-cloud render, with its hole mask as a binary confidence, is worse than no guidance at all. ARC reinforces the render it is anchored to, so it requires an aligned render and a confidence map that indicates where the render is reliable.

\begin{table}[t]
\caption{Condition quality on the unseen split. ZNCC and DINO co-location cosine measure alignment with the ground truth inside the valid circle; Sharpness is the Laplacian-variance ratio to the ground truth (1 = the recording); Time is seconds per clip on one GPU. Protocol in Appendix~\ref{app:corr-detail}.}
\label{tab:cond}
\vspace{1mm}
\centering
\footnotesize
\setlength{\tabcolsep}{5pt}
\begin{tabular}{lccccccc}
\toprule
\multirow{2}{*}{Condition} & \multicolumn{3}{c}{ZNCC$\uparrow$ (patch $p{\times}p$)} & \multirow{2}{*}{DINO$\uparrow$} & \multirow{2}{*}{Sharpness$\uparrow$} & \multirow{2}{*}{Time (s)$\downarrow$} \\
\cmidrule(lr){2-4}
 & $p{=}16$ & $32$ & $64$ & & & \\
\midrule
Point-cloud render & 0.034 & 0.047 & 0.059 & 0.476 & \textbf{0.97} & 69.2 \\
Untuned synthesizer render & 0.008 & 0.016 & 0.030 & 0.508 & 0.50 & \textbf{4.7} \\
Synthesizer render & \textbf{0.169} & \textbf{0.217} & \textbf{0.275} & \textbf{0.586} & 0.28 & \textbf{4.7} \\
\bottomrule
\end{tabular}
\end{table}

\begin{wrapfigure}{r}{0.48\linewidth}
\centering
\includegraphics[width=\linewidth]{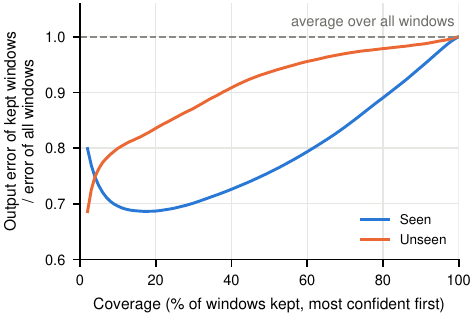}
\vspace{-20pt}
\caption{Confidence is informative. Windows of the generated video are sorted by confidence; the curve is the error of the most confident fraction relative to all windows, on Ego-Exo4D. Dashed: the average over all windows.}
\vspace{5pt}
\label{fig:gate}
\end{wrapfigure}

\vspace{-1mm}
\paragraph{Alignment versus sharpness.} We claim in Section~\ref{sec:prior} that a blurry but geometrically aligned condition is better than a sharp but misaligned one. Table~\ref{tab:cond} tests this on the unseen split by measuring both properties of each conditioning render against the ground truth. Alignment is measured by zero-mean normalized cross-correlation (ZNCC) and the similarity of DINO features \citep{simeoni2025dinov3} at the same location. Sharpness is measured by the Laplacian variance relative to the ground truth. The two renders are opposite: the synthesizer render is the blurriest but the best aligned, while the point-cloud render is sharp but poorly aligned. Despite its blur, the synthesizer render conditions the diffusion model better (Table~\ref{tab:factorial}). The blur of the condition does not remain in the output: generations from the two renders reach the same sharpness (Appendix~\ref{sec:chroma}), which indicates that the diffusion model restores the details missing from the condition. The synthesizer render is also much cheaper to compute than depth estimation and point-cloud rendering (Table~\ref{tab:cond}).

\vspace{-1mm}
\paragraph{Confidence.} Confidence gating and ARC both assume that the confidence of Eq.~\ref{eq:conf} indicates where the render is correct. Fig.~\ref{fig:gate} tests this assumption without committing to a threshold. We sort the windows of the generated videos by confidence and plot the error of the most confident windows against the ground truth as a function of how many are kept. Windows with higher confidence have lower error, well below the average over all windows at every cut-off, on seen and unseen scenes alike. The confidence therefore indicates where the render can be trusted.

\vspace{-1mm}
\paragraph{Generalization to other datasets.} Table~\ref{tab:transfer} evaluates both methods on EgoHumans \citep{khirodkar2023egohumans} and Nymeria \citep{ma2024nymeria} with the weights trained on Ego-Exo4D, without any fine-tuning. \method outperforms EgoX$^\dagger$ on both datasets on all metrics. Since $F$ is fine-tuned on Ego-Exo4D, one may suspect that it has overfit to the training environments. This is unlikely, as the new datasets contain entirely different scenes and people. The fine-tuning instead adapts $F$ to the egocentric camera model and to the large view change (Section~\ref{sec:method-camera}), which are shared across datasets. In addition, the gate keeps a similar fraction of the render on the new datasets as on unseen Ego-Exo4D scenes, roughly 15\% in each case, indicating that the synthesizer's confidence does not depend on scene familiarity. Detailed dataset construction and camera preprocessing for this transfer evaluation are provided in Appendix~\ref{app:transfer}.

\begin{figure}[t]
\centering
\includegraphics[width=\linewidth]{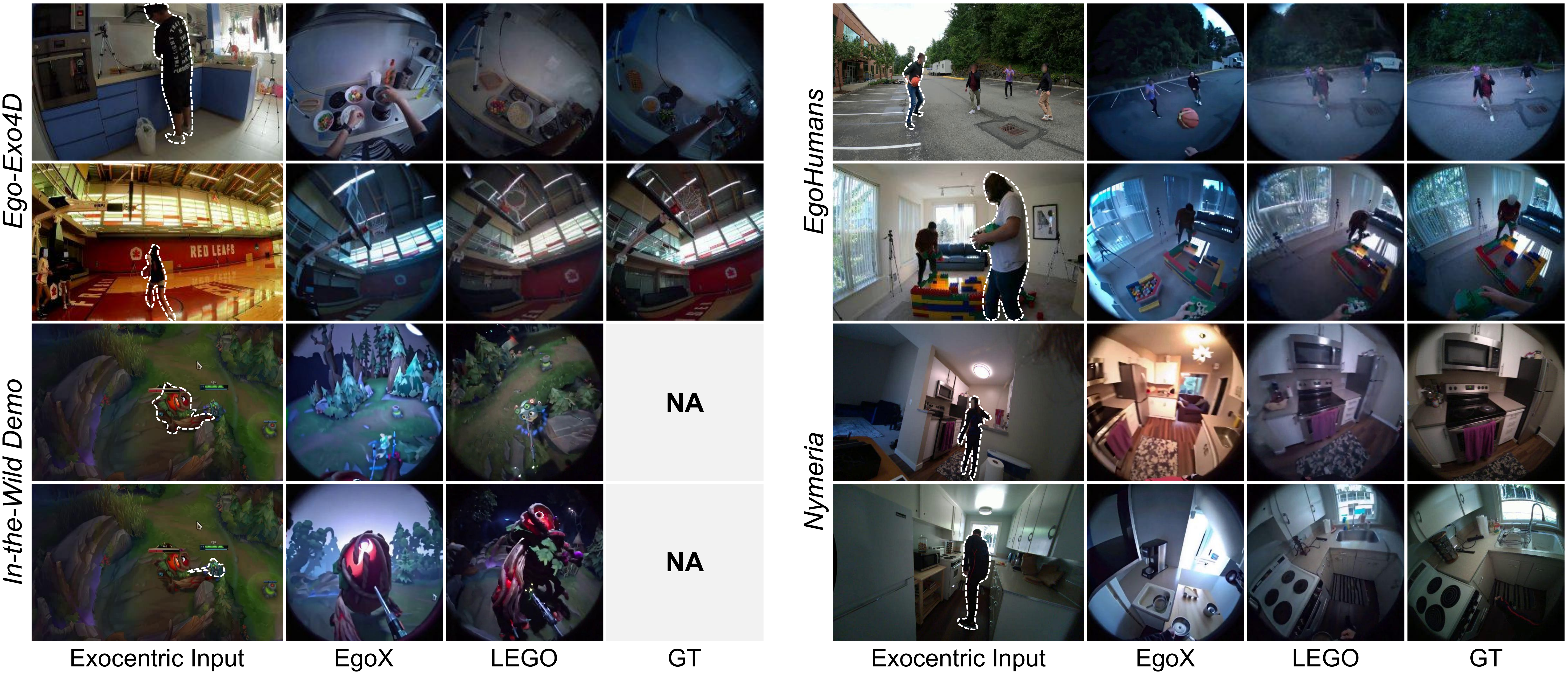}
\vspace{-20pt}
\caption{Qualitative comparison between EgoX and \method from the viewpoint indicated by the dotted outline. GT is unavailable (NA) for the in-the-wild clips. }
\label{fig:qual}
\end{figure}

\vspace{-1mm}
\paragraph{Qualitative evaluation.} Fig.~\ref{fig:qual} compares EgoX and \method on the benchmark datasets and on in-the-wild videos. EgoX typically fails where the estimated depth is wrong: the content is sharp but misplaced, most noticeably for hands and nearby objects. \method preserves the layout of the ground truth in these cases, and the diffusion model recovers the fine details that the condition lacks. On in-the-wild videos, for which no ground truth exists, the gap widens: the point-cloud render of EgoX becomes unreliable and its generated views break away from the exocentric video, whereas \method keeps them consistent. Fig.~\ref{fig:teaser} further shows \method on domains far from the training data, where it produces plausible egocentric views along user-specified camera trajectories.

\begin{table}[t]
 \caption{Evaluation on EgoHumans \citep{khirodkar2023egohumans} and Nymeria \citep{ma2024nymeria} with the weights trained on Ego-Exo4D only. Each dataset has 200 test clips. Best in bold.}
\label{tab:transfer}
\vspace{1mm}
\centering
\footnotesize
\setlength{\tabcolsep}{0pt}
\begin{tabular}{p{0.15\linewidth} p{0.12\linewidth} *{8}{>{\centering\arraybackslash}p{0.09\linewidth}}}
\toprule
& & \multicolumn{4}{c}{Image Criteria} & \multicolumn{4}{c}{Video Criteria} \\
\cmidrule(lr){3-6} \cmidrule(lr){7-10}
Dataset & Method & PSNR$\uparrow$ & SSIM$\uparrow$ & LPIPS$\downarrow$ & CLIP-I$\uparrow$ & FVD$\downarrow$ & TF$\uparrow$ & MS$\uparrow$ & DD$\uparrow$ \\
\midrule
\multirow{2}{*}{EgoHumans} & EgoX$^\dagger$ & 13.74 & 0.460 & 0.573 & 0.806 & 421.83 & 0.983 & 0.991 & \textbf{1.000} \\
& \method & \textbf{15.52} & \textbf{0.530} & \textbf{0.545} & \textbf{0.819} & \textbf{332.41} & \textbf{0.985} & \textbf{0.992} & \textbf{1.000} \\
\midrule
\multirow{2}{*}{Nymeria} & EgoX$^\dagger$ & 11.64 & 0.451 & 0.637 & 0.804 & 296.98 & 0.984 & 0.993 & \textbf{1.000} \\
& \method & \textbf{15.23} & \textbf{0.534} & \textbf{0.561} & \textbf{0.821} & \textbf{210.27} & \textbf{0.988} & \textbf{0.994} & \textbf{1.000} \\
\bottomrule
\end{tabular}
\end{table}

\section{Conclusion}
\method conditions a video diffusion model on the render of a learned view synthesizer, not a point-cloud reconstruction, on one criterion: a condition should be aligned before it is sharp, because the generator restores detail far more readily than it corrects misplaced content. The design has three practical advantages. It is lifting-free, with no depth estimation, point cloud, or reprojection at any stage. Its confidence comes free from the synthesizer's correspondence distribution and drives both the gating and a training-free guidance that replaces a bias the generator must be trained with. And the frozen synthesizer enters only through its render and confidence, so the same weights carry over to new scenes and recordings without retraining, and a stronger synthesizer can be dropped in as one appears. The synthesizer supplies the structure of the view, and the diffusion model its detail.

\subsubsection*{Reproducibility Statement}
The training and inference protocol, including every hyperparameter of the synthesizer and the generator and the fixed random seed, is given in Section~\ref{sec:exp} and Appendix~\ref{app:impl}. Appendix~\ref{app:protocol} audits our protocol parameter by parameter against the official stack of the state of the art, and Appendix~\ref{app:corr-detail} specifies the condition quality protocol. Code, configuration files, model weights, the precomputed renders and confidence maps, and the test clip lists of EgoHumans and Nymeria, sufficient to reproduce every reported configuration, will be released upon publication.

\subsubsection*{Ethics Statement}
This work uses three publicly released benchmarks, Ego-Exo4D, EgoHumans, and Nymeria, under their licenses; all were recorded with participant consent, and we collect no new data and run no human study. Generating egocentric video of a person carries privacy and impersonation risks if applied to identifiable individuals outside such consented settings. We will release our models under research-only use terms aligned with the dataset licenses, and we do not release generated videos of identifiable individuals.

\subsubsection*{The Use of Large Language Models}
Large language models were used as a writing assistant for this paper, to draft and edit text under the authors' direction. The research questions, the method, the experiments, and all reported numbers are the authors' own, and every model-assisted passage was reviewed by the authors, who take full responsibility for the content.
\bibliography{iclr2027_conference}
\bibliographystyle{iclr2027_conference}
\clearpage
\appendix
\section{Implementation Details}\label{app:impl}

\subsection{Camera Models}
The exocentric camera is treated as a pinhole with the released intrinsics. Ego-Exo4D also ships radial distortion coefficients, but they describe the lens before rectification: applying them raises the reprojection error of the annotated 3D body joints from 4 pixels to 143 pixels at $3840{\times}2160$, so we do not use them. The egocentric camera follows the Aria fisheye model, $\theta_d=\theta\,(1+\sum_i k_i\theta^{2i})$ with small tangential and thin-prism terms, which we invert per pixel by Newton iteration to obtain the target rays. All rays are expressed in the exocentric camera frame in metric units. Queries outside the image circle are not masked, so the synthesizer learns the boundary from the recordings.

\subsection{View Synthesizer}
$F$ has the architecture of the released scene-level LVSM decoder \citep{jin2025lvsm}: 24 transformer blocks of width 768 with 12 heads and patch size 8. Source tokens carry RGB and the six Pl\"ucker channels; target tokens carry the six ray channels only. We initialize from the published checkpoint and fine-tune all weights on Ego-Exo4D at $448{\times}448$ egocentric and $448{\times}784$ exocentric resolution, with an $\ell_2$ loss and an LPIPS loss weighted by 0.5. Training uses AdamW with learning rate $2{\times}10^{-5}$, cosine decay with 500 warm-up steps, a global batch size of 16, and 10k steps, about 3.5 hours on 4 H200 GPUs. The confidence of Eq.~\ref{eq:conf} is read from the attention of the last 8 layers, averaged over heads, with the softmax restricted to exocentric keys, and average-pooled $2{\times}2$ to the token grid of the generator ($28{\times}28$ egocentric queries).

\subsection{Generator}
The backbone is Wan2.1-I2V-14B at 480P \citep{wan2025wan}, fine-tuned with LoRA of rank 256 and $\alpha{=}256$ on a $49{\times}448{\times}1232$ canvas, the exocentric frame (784 wide) concatenated with the egocentric frame. Training runs for 20k steps with learning rate $2{\times}10^{-5}$, bf16, a batch size of 1 per GPU on 4 H200 GPUs, and seed 42; all ablations share this recipe. The backbone is also conditioned on a text prompt, at training and at inference alike, as in EgoX. On Ego-Exo4D we use the prompts released with the benchmark. EgoHumans and Nymeria come with none, so we generate one for every test clip from eight exocentric frames, following the system prompt of EgoX and its two-block format, an \texttt{[Exo view]} description of the scene followed by an \texttt{[Ego view]} description of the target viewpoint. We replace only the vision-language model, GPT-4o in EgoX, with the open Qwen2.5-VL \citep{bai2025qwen25vl}, and match the prompt length to the distribution of the released Ego-Exo4D prompts. At inference both methods receive the same prompt for a given clip. Inference uses 50 denoising steps with classifier-free guidance 5.0. Confidence gating uses $k{=}16$ and $\tau{=}0.3$ on the synthesizer's $56{\times}56$ query grid. ARC applies Eq.~\ref{eq:arc} after each denoising step during the first 80\% of the schedule ($h{=}0.8$), with the noise level floored at 0.05. The ARC hyperparameters ($h$ and the squared weighting) were chosen on the unseen split, since Ego-Exo4D provides no separate validation split.

\section{Evaluation Details}\label{app:eval}

\subsection{Comparison Protocol}\label{app:protocol}
To compare fairly with EgoX \citep{kang2026egox}, we match its training and inference configuration wherever the released code specifies it: backbone checkpoint, LoRA rank and alpha, canvas resolution, optimizer, learning rate, precision, batch size, seed, noising and loss masks, sampler, number of denoising steps, guidance scale, and frame count. The GGA ablation uses the official implementation of the geometry-guided attention bias, and we verified that the bias is applied identically at training and inference. Two differences remain. We train for 20k steps with a batch size of 1 per GPU on 4 H200 GPUs for about four days, whereas EgoX trains with the same batch size on 8 H200 GPUs for about one day. Second, since the evaluation code is not public, metrics are our own implementation. Under our implementation (Appendix~\ref{app:metrics}), the released checkpoint scores 15.57 (16.05) dB PSNR on seen and 13.53 (14.38) dB on unseen scenes, where the numbers in parentheses are those reported in \citet{kang2026egox}.

\subsection{Image and Video Metrics}\label{app:metrics}
Since the evaluation code of EgoX \citep{kang2026egox} is not publicly available, we reimplemented all metrics following the descriptions in the paper and calibrated our implementation so that the released EgoX checkpoint approximately reproduces the published numbers (Appendix~\ref{app:protocol}). Generated and ground-truth videos are compared over the 49 frames of each clip; frame scores are averaged within a clip and clip scores over the split.

\vspace{-1mm}
\paragraph{Image criteria.} PSNR, SSIM, LPIPS, and CLIP-I are computed on the $448{\times}448$ egocentric half of the $1232{\times}448$ canvas. SSIM uses an $11{\times}11$ Gaussian window with $\sigma{=}1.5$. LPIPS \citep{zhang2018lpips} uses the AlexNet backbone, and CLIP-I is the cosine similarity between CLIP ViT-B/32 \citep{radford2021clip} embeddings of the generated and ground-truth frames.

\vspace{-1mm}
\paragraph{Video criteria.} FVD \citep{unterthiner2018fvd} is computed with the \texttt{cd-fvd} implementation \citep{ge2024fvdbias} on the egocentric half: I3D features of one 16-frame sample per video, resized to $224{\times}224$, compared between the generated and ground-truth clips of a split. Temporal flickering, motion smoothness, and dynamic degree are computed with the official VBench code \citep{huang2024vbench} on the full canvas, which reproduces the values reported by EgoX (Appendix~\ref{app:protocol}). Dynamic degree is a per-clip binary decision based on RAFT optical flow; we keep the VBench sampling and count rule and apply the same flow threshold to every method.

\vspace{-1mm}
\paragraph{Sharpness and chroma.} For the analysis of the conditioning renders, sharpness is the variance of the Laplacian response of a frame on a region $\Omega$, reported as the ratio to the ground-truth frame on the same region. $\Omega$ is the valid egocentric image circle, or, for point-cloud renders, the rendered pixels eroded by 3\,px. Chroma is the mean chroma magnitude inside the valid image circle, reported as the ratio of the generated to the ground-truth video. Both ratios are averaged over the 13 frames aligned with the latent frames ($t{=}0,4,\ldots,48$) and over clips; a value of 1 means the same amount as the ground truth.

\subsection{Object-Level Metrics}\label{app:object}
EgoX \citep{kang2026egox} reports three object-level metrics: \emph{LocErr}, the distance between the bounding-box centers of matched objects in the generated and ground-truth frames; \emph{IoU}, the overlap of their bounding boxes; and \emph{Contour}, the agreement of their shapes. Computing them requires a detection and matching pipeline that is not part of the public release. We therefore fix one protocol, validate it against the published numbers, and apply it unchanged to all methods.

\vspace{-1mm}
\paragraph{Protocol.} Objects are segmented and tracked with SAM~2 \citep{ravi2024sam2} on the generated and the ground-truth video independently, using automatic mask generation on the first frame (32 points per side, seed boxes of at least 32\,px, mask regions of at least 2000\,px$^2$). Each object crop is embedded with DINOv3 ViT-L \citep{simeoni2025dinov3}, and two objects in the same frame are matched when the cosine similarity of their embeddings is at least 0.9, without one-to-one assignment, as described in EgoX. LocErr and IoU are computed on the matched boxes and averaged per clip and then over clips. For Contour, the literal definition in EgoX is the mask IoU in image coordinates, but it cannot reproduce the published values, which exceed the published box IoU by a wide margin; the boundary F-score after aligning the mask centroids does reproduce them, so we adopt it.

\vspace{-1mm}
\paragraph{Calibration.} EgoX defines the three metrics but not the segmentation, the object features, or the matching threshold used to compute them. We chose these settings by searching over the corresponding options and scoring each combination by the relative error of the released EgoX checkpoint against the six published numbers. Combinations that retain fewer than 30 matched pairs on either split are discarded.

\vspace{-1mm}
\paragraph{Validation.} Under the protocol above, the released checkpoint reproduces the published unseen row almost exactly (LocErr 146.5 against 149.9, IoU 0.091 against 0.092), and \method outperforms it on all three metrics on both splits (Table~\ref{tab:object}). The published seen row is not reproduced by any combination in the search: the closest one still leaves a large gap for the released checkpoint (LocErr 87.5 against 61.8, IoU 0.222 against 0.363) and no longer agrees on the unseen row. We therefore quote the published seen row for reference only. The lower block of Table~\ref{tab:object} scores both methods under that closest combination, a permissive segmenter with DINOv3-B patch features, one-to-one matching at $\tau{=}0.56$, and registered mask-IoU contour, where \method reaches the level of the published seen row while the released checkpoint does not.
\begin{table}[t]
\caption{Object-level criteria under two protocols. \emph{Top}: our fixed protocol, which reproduces the published unseen row from the released checkpoint. \emph{Bottom}: the sweep combination closest to the published seen row. Published numbers are \quoted{quoted}.}
\label{tab:object}
\vspace{1mm}
\centering
\footnotesize
\setlength{\tabcolsep}{2pt}
\begin{tabular}{>{\raggedright\arraybackslash}p{0.34\linewidth} *{3}{c} *{3}{c}}
\toprule
& \multicolumn{3}{c}{Seen} & \multicolumn{3}{c}{Unseen} \\
\cmidrule(lr){2-4} \cmidrule(lr){5-7}
System & LocErr$\downarrow$ & IoU$\uparrow$ & Contour$\uparrow$ & LocErr$\downarrow$ & IoU$\uparrow$ & Contour$\uparrow$ \\
\midrule
\multicolumn{7}{l}{\textit{Fixed protocol (ours; validated on the published unseen row)}} \\
EgoX (published) & \quoted{61.8} & \quoted{0.363} & \quoted{0.546} & \quoted{149.9} & \quoted{0.092} & \quoted{0.481} \\
EgoX (released)$^\dagger$ & 125.3 & 0.176 & 0.622 & 146.5 & 0.091 & 0.572 \\
\method & 107.7 & 0.279 & 0.674 & 143.3 & 0.097 & 0.609 \\
\addlinespace[3pt]
\multicolumn{7}{l}{\textit{Seen-fitted protocol (closest sweep combination to the published seen row)}} \\
EgoX (released)$^\dagger$ & 87.5 & 0.222 & 0.546 & 115.4 & 0.119 & 0.477 \\
\method & 65.3 & 0.434 & 0.634 & 116.5 & 0.124 & 0.488 \\
\bottomrule
\end{tabular}
\end{table}

\subsection{Condition Quality Metrics}\label{app:corr-detail}
Table~\ref{tab:cond} compares each conditioning render against the ground-truth egocentric frame on the unseen split, with metrics that depend on neither the synthesizer nor the diffusion model. We evaluate the 13 frames that align with the latent frames of the generator ($t{=}0,4,\dots,48$).

\vspace{-1mm}
\paragraph{Evaluation region.} The valid circle of a clip is the set of pixels that are lit in any ground-truth frame, eroded by 4\,px. For the point-cloud render the region is further restricted to the rendered pixels, eroded by 3\,px so that the edges of the holes are excluded.

\vspace{-1mm}
\paragraph{Metrics.} ZNCC is the zero-mean normalized cross-correlation of luminance, computed over non-overlapping $p{\times}p$ patches in which at least half of the pixels are valid, and averaged over patches, frames, and clips. DINO co-location is the cosine similarity between the DINOv3 ViT-B/16 patch tokens of the render and of the ground truth at the same location, averaged over tokens whose $16{\times}16$ window is at least half valid. Sharpness is the ratio of the Laplacian variance of the render to that of the ground truth on the same pixels, and coverage is the fraction of the valid circle that the render fills. Alignment in dB is the PSNR of the render against the ground truth on the same evaluation region.

\subsection{Cross-Dataset Evaluation}\label{app:transfer}
This subsection describes how the test clips of EgoHumans and Nymeria are selected and how their videos are prepared for evaluation. Both datasets record the egocentric view with Project Aria glasses, so the egocentric camera is the same fisheye model as in Ego-Exo4D.

\vspace{-1mm}
\paragraph{EgoHumans.} This dataset \citep{khirodkar2023egohumans} records multi-person activities with several exocentric cameras and is not used for training. We select 200 clips of 49 frames, distributed evenly over the seven activity scenarios and drawn from multiple takes per scenario. Within a take, the wearer and the starting frame are sampled at regular intervals with a fixed seed. For each clip we use a single exocentric camera, the one in which the wearer is closest to the image center. The exocentric frames are resized and center-cropped to the canvas resolution, and the egocentric frames are the raw fisheye recordings, as in Ego-Exo4D.

\vspace{-1mm}
\paragraph{Nymeria.} This dataset \citep{ma2024nymeria} records a participant and an observer, both wearing Aria glasses, and is not used for training. The observer takes the role of the exocentric camera. Since the observer moves, we use only 49-frame windows in which the observer camera is effectively static, with at most 4\,cm of translation and $2.5^\circ$ of rotation. We further require that the participant's head is visible in the observer's view throughout the window, which we check by projecting the participant's camera center into the observer's frame and by detecting a person at the projected location. From the resulting pool we select 200 clips over 26 sequences, balancing the number of clips per sequence. The observer's fisheye frames are undistorted to a pinhole camera with a focal length of 260 pixels and cropped to the canvas resolution.

\vspace{-1mm}
\paragraph{Evaluation.} Both datasets are scored with the evaluation code on the egocentric frames of the same clips. EgoX is run with the authors' released checkpoint and pipeline, including depth estimation, point-cloud rendering, and geometry-guided attention. Both methods use the same hyperparameters as on Ego-Exo4D.

\section{Discussions}\label{app:analysis}

\subsection{Alignment, Sharpness, and Coverage}\label{sec:sharpalign}

The synthesizer render is better aligned than the point-cloud render by $4.2$ dB inside the valid circle. This advantage remains even when we evaluate alignment only on pixels where the point-cloud render is valid, indicating that the difference is not simply due to the point cloud's missing regions.

High coverage also does not guarantee good alignment. The untuned synthesizer covers $99.6\%$ of the valid circle and is sharper than the fine-tuned synthesizer ($0.50$ vs.\ $0.28$), yet it has the worst ZNCC alignment of the three. The point-cloud render is locally sharp where geometry is available, but occlusion and depth errors reduce its coverage to $53.4\%$, compared with $98.1\%$ for the synthesizer render.

\subsection{Output Sharpness and Color}\label{sec:chroma}
Despite the difference in sharpness between the conditioning renders, their generated outputs have similar sharpness. On unseen scenes, the Laplacian-variance ratio to the ground truth is $1.10$ for outputs conditioned on the point-cloud render and $1.04$ for those conditioned on the synthesizer render. This indicates that output sharpness is not strongly determined by the sharpness of the conditioning render.

Color behaves differently. On unseen scenes, the synthesizer render has a chroma ratio of only $0.41$ relative to the ground truth, and the generated output shows a similar reduction, reaching $0.45$. In comparison, the output reaches $0.82$ when conditioned on the point-cloud render and $1.26$ without conditioning. Gating the synthesizer render partially recovers the chroma ratio to $0.62$, suggesting that removing unreliable conditioning allows the generator to rely more on its own appearance prior. Thus, unlike sharpness, output color remains more sensitive to the conditioning render.

The mechanism is an asymmetry between the two models. The synthesizer is a regression model, so under correspondence uncertainty it tends toward a conditional mean, in which chroma averages toward gray while luminance structure largely survives. The generator under the same uncertainty samples a plausible mode, and its color prior, acquired at pretraining scale and adapted here only through a low-rank update, is unaffected by the synthesizer's collapse. Such conditioning over-reliance is documented for image-to-video models \citep{zhao2024leakage}. The standard remedy, for it and for the exposure bias, is to degrade or noise-augment the conditioning during training \citep{ho2022cascaded,blattmann2023svd}. Color drift is itself a known signature of conditioning on model-produced imagery \citep{li2026freqforcing}. FVD is particularly sensitive to such per-frame appearance statistics \citep{ge2024fvdbias}, which accounts for part of the unseen FVD gap in Table~\ref{tab:factorial}.

\subsection{Adapting to the Fisheye Camera}\label{sec:spill}
The untuned synthesizer does not reliably respect the target camera model, even when given the correct camera rays. On unseen scenes, it fills $95.9\%$ of the region outside the valid fisheye circle with scene content. We measure this behavior using \emph{spill}, defined as the fraction of the region outside the image circle that a signal fills with content.

Simply masking this region would not solve the problem. Inside the valid field of view, the untuned synthesizer is also the worst aligned of the three conditions, trailing the fine-tuned synthesizer by $5.4$ dB on unseen scenes and $10.3$ dB on seen scenes. Conditioning the generator on this render further reduces performance by $0.8$ dB on unseen scenes and $2.5$ dB on seen scenes.

Fine-tuning substantially corrects both behaviors. The adapted synthesizer leaves the region outside the image circle essentially empty, with only $0.1\%$ spill, while the point-cloud render fills $22.4\%$. Adapting the synthesizer to the target camera is therefore necessary, not optional.

\begin{figure}[t]
\centering
\includegraphics[width=\linewidth]{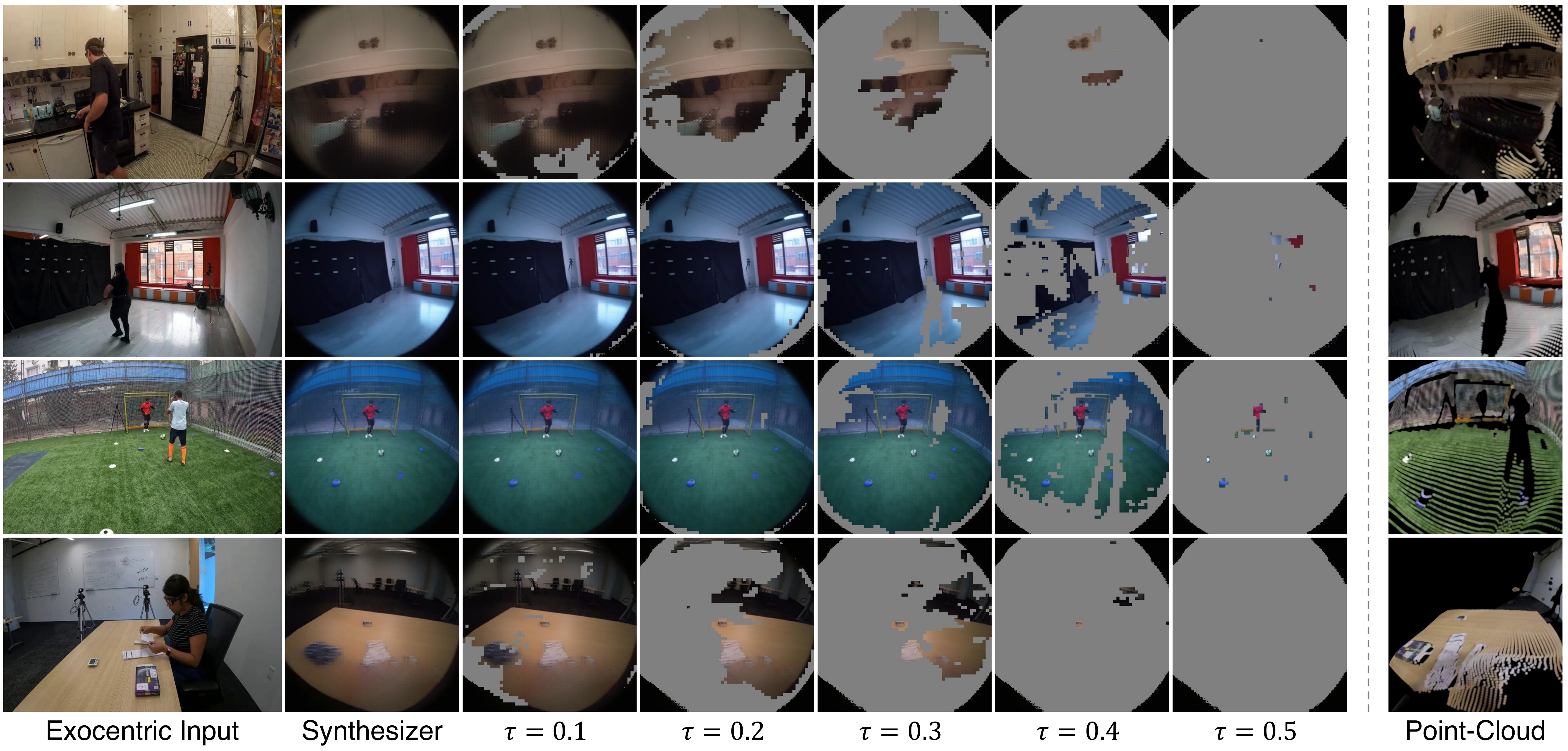}
\vspace{-20pt}
\caption{Effect of the gate threshold. The synthesizer render gated at
$\tau=0.1,\dots,0.5$, shown between the ungated render and the point-cloud
render of \citet{kang2026egox}.}
\label{fig:supp_gating}
\end{figure}

\subsection{Confidence Gating}\label{app:abstain}

The gate of Eq.~\ref{eq:conf} restores abstention from the same distribution that produced the render. The confidence is the probability mass of the top $k{=}16$ sources, read from the last $8$ layers with head averaging. Tokens below $\tau{=}0.3$ are set to the neutral gray that the VAE maps near zero, which is the same value as an empty conditioning latent (Fig.~\ref{fig:supp_gating}).

\vspace{-1mm}
\paragraph{Gate statistics (Fig.~\ref{fig:gate}).}
To test whether confidence tracks render reliability independently of the threshold, we sort $16{\times}16$ windows by their mean confidence and plot the luminance error of the most confident fraction. We report errors as $(\text{kept}, \text{discarded})$. At $\tau{=}0.3$, render errors are $(11.6, 16.6)$ on seen scenes and $(25.9, 30.7)$ on unseen scenes; output errors show the same trend, at $(13.1, 21.4)$ and $(26.1, 32.8)$, respectively. On both splits, kept windows consistently have lower render and output errors than discarded windows, indicating that the confidence score tracks both rendering reliability and final generation quality. The render curves have the same shape as the output curves shown in Fig.~\ref{fig:gate}. We report absolute luminance error rather than ZNCC because window-level ZNCC is confounded by texture; within strata of ground-truth texture, kept windows still have higher ZNCC than discarded ones.

\subsection{Computational Cost}\label{app:runtime}
\begin{table}[t]
\centering
\footnotesize
\setlength{\tabcolsep}{4pt}
\caption{Cost of generating one clip on one H200 GPU, excluding the one-time model load. Generation assumes that the condition is already built, and end-to-end includes it. Best in bold.}
\label{tab:runtime}
\vspace{1mm}
\begin{tabular}{p{0.23\linewidth} cc}
\toprule
& EgoX$^\dagger$ & \method \\
\midrule
Denoiser bias & GGA & ARC \\
\midrule
Denoising loop (s)$\downarrow$ & 565.3 & \textbf{540.3} \\
Generation (s)$\downarrow$ & 597.5 & \textbf{573.2} \\
\midrule
End-to-end (s)$\downarrow$ & 666.7 & \textbf{577.9} \\
Peak VRAM (GiB)$\downarrow$ & 69.9 & \textbf{68.2} \\
\bottomrule
\end{tabular}
\end{table}

Table~\ref{tab:runtime} reports the cost of generating one clip on a single H200 GPU, averaged over the 100 clips. Generation covers the denoising loop together with the encoding and decoding around it, and assumes that the condition is already built. End-to-end adds the cost of building the condition. During denoising, the two methods differ only in the bias they apply. GGA builds a dense attention bias over the whole canvas and reapplies it in every transformer block, which adds 252\,ms to each forward pass. ARC instead updates the egocentric part of the clean estimate once per step, in 2.7\,ms. The larger difference is upstream. Building our condition takes 4.7\,s per clip, against 69.2\,s for depth estimation and point-cloud rendering (Table~\ref{tab:cond}). Overall, \method generates a clip faster than EgoX and uses slightly less memory.

\subsection{Limitations}\label{app:limit}
\begin{figure}[t]
\centering
\includegraphics[width=0.85\linewidth]{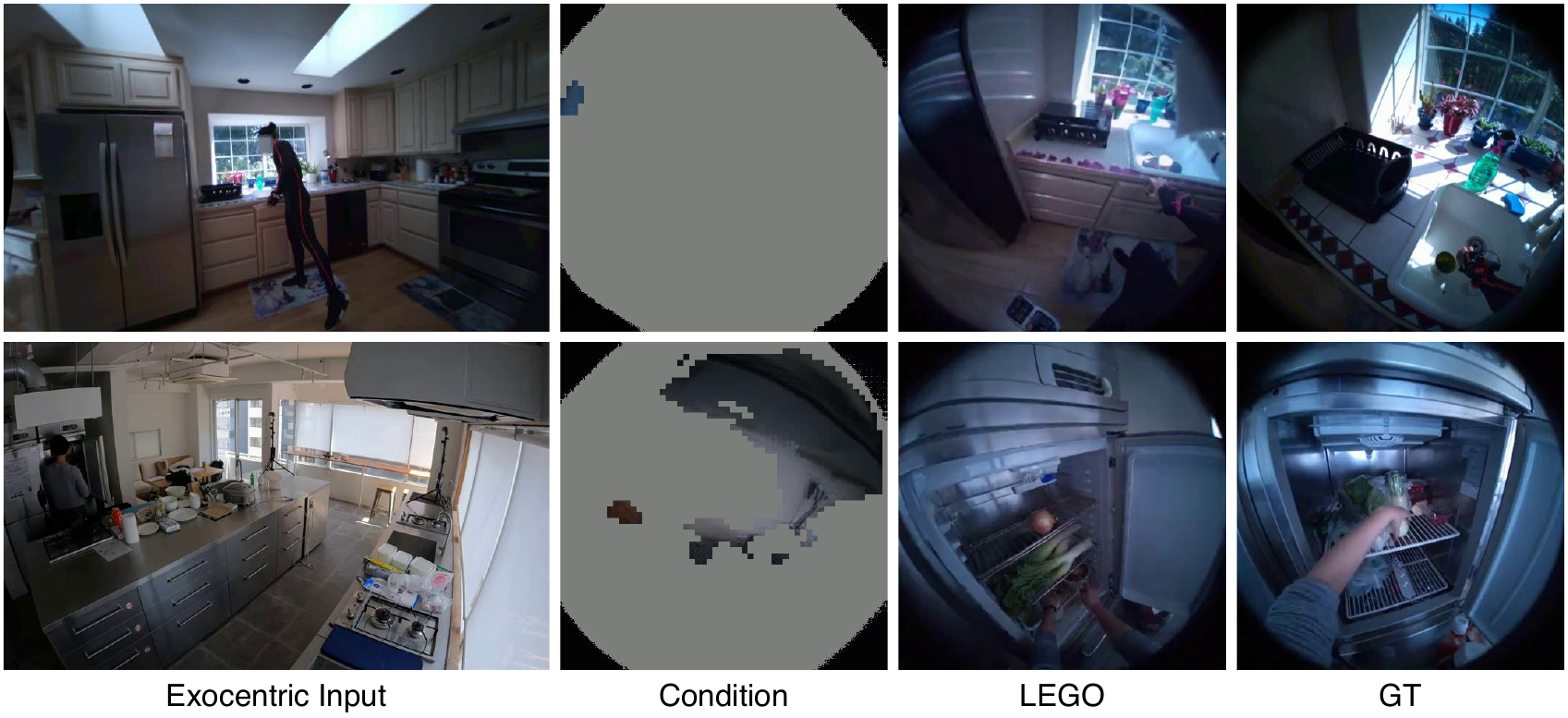}
\vspace{-5pt}
\caption{A limitation from $F$. Where the exocentric frame does not cover the egocentric view, the condition is empty and the generator fills the region from its own prior.}
\label{fig:supp_limitation}
\end{figure}

\method inherits the limitations of the view synthesizer. $F$ renders each frame independently from a single exocentric frame, so it cannot recover content that the frame does not show (Fig.~\ref{fig:supp_limitation}), and it blurs where the correspondence is uncertain. Its colors also collapse on unfamiliar scenes (Section~\ref{sec:chroma}). The pipeline is lifting-free but still requires the camera poses of both views. Since $F$ is frozen and enters only through the conditioning input and the confidence map, a stronger feed-forward view synthesizer can replace it without changing the diffusion model or ARC.

\section{Additional Qualitative Results}\label{app:qual}
\begin{figure}[t]
\centering
\includegraphics[width=0.85\linewidth]{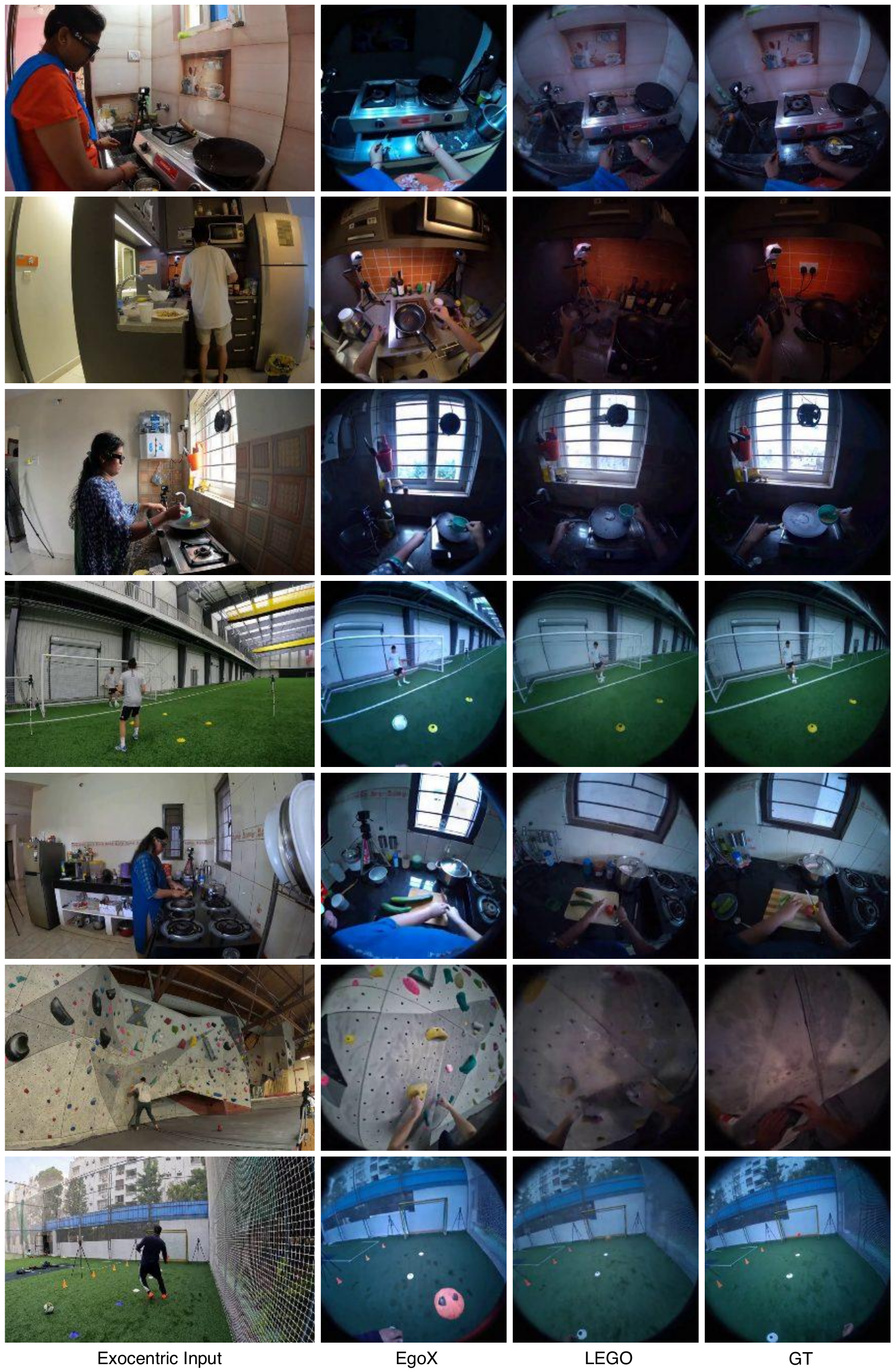}
\vspace{-5pt}
\caption{Additional qualitative comparisons on the Ego-Exo4D seen split.}
\label{fig:qual_supp_seen}
\end{figure}

\begin{figure}[t]
\centering
\includegraphics[width=0.85\linewidth]{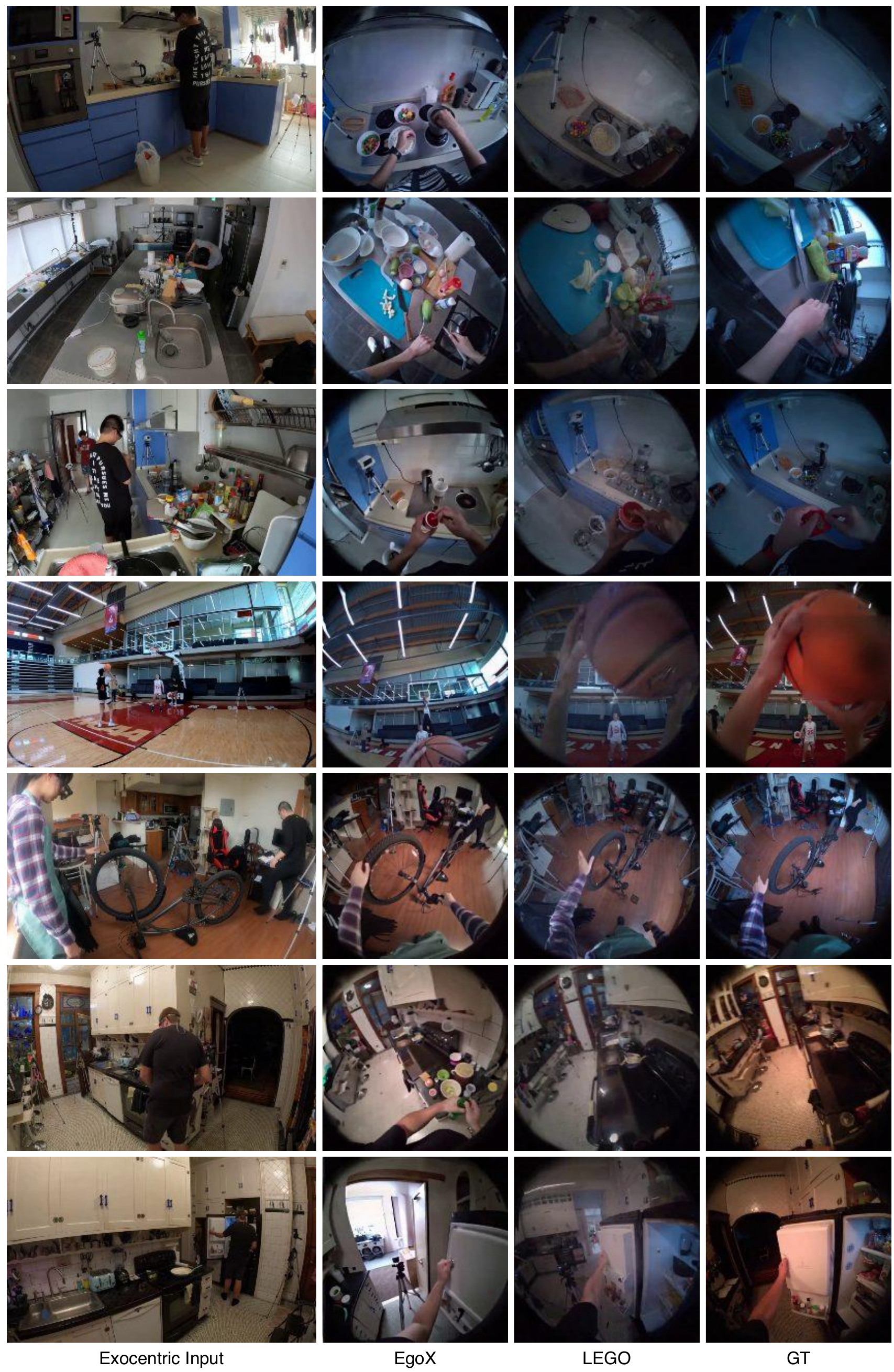}
\vspace{-5pt}
\caption{Additional qualitative comparisons on the Ego-Exo4D unseen split.}
\label{fig:qual_supp_unseen}
\end{figure}

\begin{figure}[t]
\centering
\includegraphics[width=0.85\linewidth]{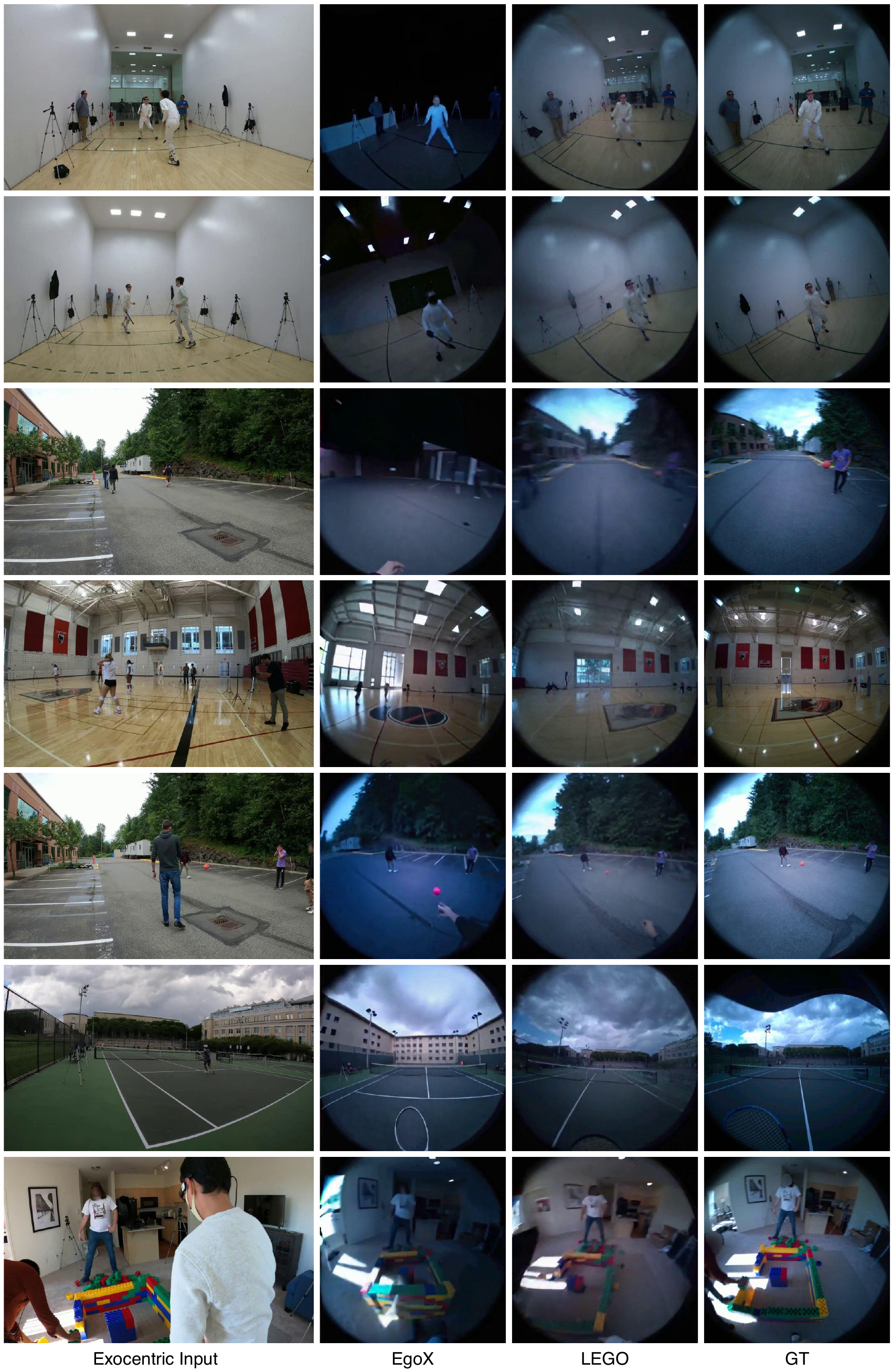}
\vspace{-5pt}
\caption{Additional qualitative comparisons on EgoHumans.}
\label{fig:qual_supp_egohumans}
\end{figure}

\begin{figure}[t]
\centering
\includegraphics[width=0.85\linewidth]{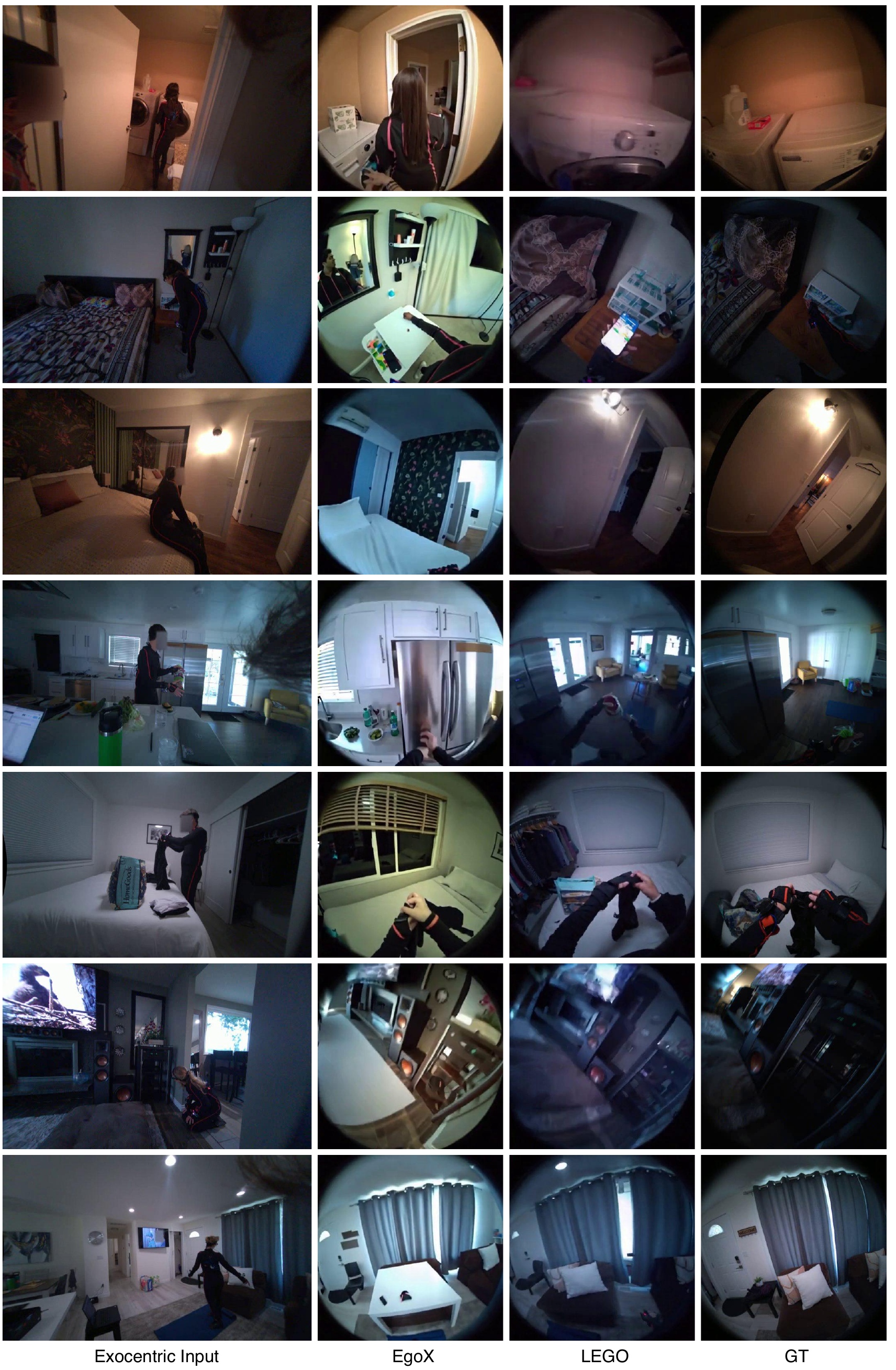}
\vspace{-5pt}
\caption{Additional qualitative comparisons on Nymeria.}
\label{fig:qual_supp_nymeria}
\end{figure}

Figures~\ref{fig:qual_supp_seen} and~\ref{fig:qual_supp_unseen} show further comparisons on the seen and unseen splits of Ego-Exo4D, and Figures~\ref{fig:qual_supp_egohumans} and~\ref{fig:qual_supp_nymeria} on EgoHumans and Nymeria. Each method runs its own checkpoint, the released one for EgoX and ours for \method, and both receive the same camera poses, the same text prompt, and the same random seed.

Across the examples, the two methods fail in the way described in Section~\ref{sec:exp}. EgoX reproduces content sharply at the wrong position, which is most visible for the hands and for objects close to the camera. \method keeps the scene where the ground truth places it and lets the diffusion model add the details that its condition does not carry. The same holds on EgoHumans and Nymeria. The remaining errors of \method are softer details and, on unseen scenes, less saturated colors (Section~\ref{sec:chroma}).

\newpage

\end{document}